\documentclass[10pt, letterpaper, journal, twoside]{IEEEtran}
\IEEEoverridecommandlockouts    
   
\usepackage[letterpaper,top=0.70in,bottom=0.59in,left=0.60in,right=0.60in]{geometry}

\usepackage{graphics} 
\usepackage{cite}
\usepackage[T1]{fontenc}
\usepackage{amsmath, bm} 
\usepackage{cancel} 
\usepackage{amssymb}  
\usepackage{hyperref}
\hypersetup{
    colorlinks=true,
    linkcolor=blue,
    filecolor=magenta,      
    urlcolor=blue,
}
\usepackage{comment}    
\usepackage[normalem]{ulem} 

\newif\ifhideequation
\hideequationfalse
\ifhideequation
  \excludecomment{eqblock}
\else
  \includecomment{eqblock}
\fi

\usepackage{float}
\usepackage{scalerel}
\newcommand\Tau{\footnotesize \scalerel*{\tau}{T}}

\usepackage{numprint}
\usepackage{comment}

\usepackage[T1]{fontenc}
\usepackage{pgfplots}
\usepackage{amssymb}
\pgfplotsset{compat=newest}
\usetikzlibrary{plotmarks}
\usetikzlibrary{arrows.meta}
\usepgfplotslibrary{patchplots}
\usepackage{amsmath}

\usepackage{color, colortbl}
\definecolor{LightCyan}{rgb}{0.88,1,1}
\usepackage{multirow}
\usepackage{siunitx}
\DeclareSIUnit\cell{cell}
\DeclareSIUnit\cells{cells}
\DeclareSIUnit\trees{trees}

\newcommand\highlightReference[1]{%
  \expandafter\newcommand\csname highlightReference-#1\endcsname{}%
}
\let\oldbibitem\bibitem
\def\bibitem#1#2\par{%
  \expandafter\ifx\csname highlightReference-#1\endcsname\relax
    \oldbibitem{#1}#2\par
  \else
    \oldbibitem{#1}\highlight{#2}\par
  \fi
}
\usepackage{color}
\newcommand\highlight[1]{\textcolor{red}{#1}}
\newcommand\highlightStrike[1]{\textcolor{red}{\sout{#1}}}
\newcommand\strikeReference[1]{%
  \expandafter\newcommand\csname strikeReference-#1\endcsname{}%
}
\def\bibitem#1#2\par{%
  \expandafter\ifx\csname strikeReference-#1\endcsname\relax
    \expandafter\ifx\csname highlightReference-#1\endcsname\relax
      \oldbibitem{#1}#2\par
    \else
      \oldbibitem{#1}\highlight{#2}\par
    \fi
  \else
    \oldbibitem{#1}\highlightStrike{#2}\par
  \fi
}
\usepackage{titlesec} 
\renewcommand{\thesection}{\Roman{section}}
\renewcommand{\thesubsection}{\thesection-\Alph{subsection}}
\renewcommand{\thesubsubsection}{\thesubsection\arabic{subsubsection}}
\titleformat{\subsubsection}[runin]{\itshape}{\arabic{subsubsection})}{0.5em}{}

\titlespacing*{\section}
{0pt}{2pt}{2pt} 
\titlespacing*{\subsection}
{0pt}{2pt}{2pt} 
\titlespacing*{\subsubsection}
{8pt}{1pt}{2pt} 
\usepackage{enumitem} 

\usepackage{wrapfig}
\usepackage{amssymb}
\usepackage{tikz}
\usetikzlibrary{shapes,arrows}
\usepackage{verbatim}
\usetikzlibrary{positioning}
\usepackage{tabstackengine}
\usepackage{algorithm}
\usepackage[end]{algpseudocode} 
\usepackage{etoolbox}
\usepackage[font=small, skip=1pt]{caption}
\usepackage[skip=-1pt]{subcaption} 

\usepackage[nodisplayskipstretch]{setspace}
\usepackage{pgfplots}
\usepackage{grffile}
\pgfplotsset{compat=newest}
\usetikzlibrary{plotmarks}
\usetikzlibrary{arrows.meta}
\usepgfplotslibrary{patchplots}
\usepackage{amsmath}

\usepackage{bbding} 

\usepackage{colortbl}

\usepackage{hhline}
\usepackage{makecell}

\newcommand{\ihab}[1]{{\textcolor{black}{#1}}}

\usepackage{eso-pic}
\newcommand\AtPageUpperMyright[1]{\AtPageUpperLeft{%
 \put(\LenToUnit{0.5\paperwidth},\LenToUnit{-1.5cm}){%
     \parbox{0.5\textwidth}{\raggedright\fontsize{11}{11}\selectfont #1}}%
 }}%
\newcommand{\conf}[1]{%
\AddToShipoutPictureBG*{%
\AtPageUpperMyright{#1}
}
}

\title{\LARGE \bf 
Chance-Constrained Sampling-Based MPC for Collision Avoidance in Uncertain Dynamic Environments
}

\conf{\hspace*{-9cm}\textcolor{gray}{This paper has been accepted for publication in \textbf{IEEE Robotics and Automation Letters} \textcolor{blue}{\href{https://ieeexplore.ieee.org/xpl/RecentIssue.jsp?punumber=7083369}{(RA-L)}}, May 2025.}}

\author{Ihab S. Mohamed, Mahmoud Ali, and Lantao Liu
\thanks{Authors are with the Luddy School of Informatics, Computing, and Engineering, Indiana University, Bloomington, IN 47408 USA (e-mail: {\tt\small \{mohamedi, alimaa, lantao\}@iu.edu}). \\ 
This work was supported by the National Science Foundation under Grant No. 2047169.
 }

}%
\definecolor{applegreen}{rgb}{0.8, 1, 0.0}
\definecolor{LightCyan}{rgb}{0.88,1,1}
\definecolor{atomictangerine}{rgb}{1.0, 0.6, 0.4}
\definecolor{amber}{rgb}{1.0, 0.75, 0.0}
\definecolor{aqua}{rgb}{0.0, 1.0, 1.0}
\definecolor{almond}{rgb}{0.94, 0.87, 0.8}
\definecolor{aquamarine}{rgb}{0.5, 1.0, 0.83}
\definecolor{babyblue}{rgb}{0.54, 0.81, 0.94}
\definecolor{babyblueeyes}{rgb}{0.63, 0.79, 0.95}
\definecolor{asparagus}{rgb}{0.53, 0.66, 0.42}
\definecolor{auburn}{rgb}{0.43, 0.21, 0.1}
\definecolor{brilliantlavender}{rgb}{0.96, 0.73, 1.0}
\definecolor{bittersweet}{rgb}{1.0, 0.44, 0.37}
\definecolor{blue-violet}{rgb}{0.54, 0.17, 0.89}
\definecolor{capri}{rgb}{0.0, 0.75, 1.0}
\definecolor{celadon}{rgb}{0.67, 0.88, 0.69}
\definecolor{darkcyan}{rgb}{0.0, 0.55, 0.55}
\definecolor{deepskyblue}{rgb}{0.0, 0.75, 1.0}
\definecolor{dogwoodrose}{rgb}{0.84, 0.09, 0.41}

\begin{document}

\maketitle

\global\csname @topnum\endcsname 0
\global\csname @botnum\endcsname 0


\thispagestyle{empty}
\pagestyle{empty}


\vspace*{-27pt}
\begin{abstract}
Navigating safely in dynamic and uncertain environments is challenging due to uncertainties in perception and motion.
This letter presents the Chance-Constrained Unscented Model Predictive Path Integral (C\textsuperscript{2}U-MPPI) framework, a robust sampling-based Model Predictive Control (MPC) algorithm that addresses these challenges by leveraging the U-MPPI control strategy with integrated probabilistic chance constraints, enabling more reliable and efficient navigation under uncertainty.
Unlike gradient-based MPC methods, our approach (i) avoids linearization of system dynamics by directly applying non-convex and nonlinear chance constraints, enabling more accurate and flexible optimization, and (ii) enhances computational efficiency by leveraging a deterministic form of probabilistic constraints and employing a layered dynamic obstacle representation, enabling real-time handling of multiple obstacles.
Extensive experiments in simulated and real-world human-shared environments validate the effectiveness of our algorithm against baseline methods, showcasing its capability to generate feasible trajectories and control inputs that adhere to system dynamics and constraints in dynamic settings, enabled by unscented-based sampling strategy and risk-sensitive trajectory evaluation.
A supplementary video is available at:~\url{https://youtu.be/FptAhvJlQm8}.

\end{abstract}
\vspace*{-3pt}
\begin{IEEEkeywords}
Dense crowds, robot navigation, collision avoidance, optimization and optimal control 
\end{IEEEkeywords}
\vspace*{-2pt}
\section{Introduction and Related Work}\label{Introduction}
\begin{figure}%
    \vspace*{-37pt}
    \centering
    \subfloat[LKF-based predictions]{%
        \vspace*{-6pt}
        \includegraphics[scale=0.66]{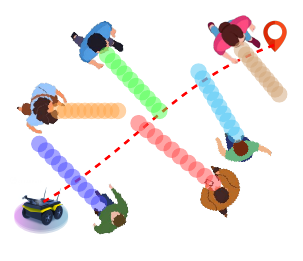}%
        \label{fig:person-predictions}}%
    \hspace{10pt}
    \subfloat[Layered representation]{%
        \vspace*{-10pt}
        \includegraphics[scale=0.93]{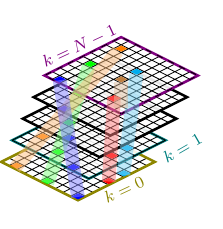}%
        \label{fig:layered-obstacle}}%
    \vspace*{-5pt}
    \caption{Handling dynamic obstacles by (a) predicting pedestrian positions using LKF and (b) structuring them into a layered representation for efficient trajectory evaluation over the prediction horizon \( N \).}
    \label{fig:predictions-representation}%
    \vspace*{-13pt}
\end{figure}

Nowadays, autonomous ground vehicles (AGVs) are increasingly deployed in complex, human-shared environments such as smart warehouses, autonomous driving, and urban logistics hubs. In these settings, AGVs must navigate toward designated goals while avoiding both static and dynamic obstacles.
This task is particularly challenging due to uncertainties in predicting the future states of moving obstacles and potential localization errors, which complicate safety assessments of planned trajectories. As a result, AGVs must effectively balance safety and efficiency, underscoring the importance of \textit{risk-aware} motion planning and control in unstructured, uncertain, and dynamic environments~\cite{singamaneni2024survey}.

Numerous collision avoidance techniques have been developed to quantify and mitigate risks in dynamic environments, broadly classified into three main categories: reactive-based, optimization-based, and learning-based approaches~\cite{ferrer2013robot, khatib1986real, chen2017socially, brito2019model, chen2021interactive, stefanini2024efficient}.
Reactive-based methods, which rely on specific rules for collision avoidance, such as social forces~\cite{ferrer2013robot} and artificial potential fields~\cite{khatib1986real}, are computationally efficient but often fail to capture human behavior accurately, sometimes resulting in unsafe or unnatural movements. In contrast, learning-based approaches aim to emulate and anticipate human behavior but encounter challenges in unfamiliar scenarios due to their dependence on prior training data~\cite{chen2017socially}. Furthermore, both approaches typically neglect the robot’s kinodynamic constraints, limiting their effectiveness in complex or high-speed scenarios. 
Optimization-based methods, particularly Model Predictive Control (MPC), have emerged as prominent solutions to address these limitations, effectively integrating the robot kinodynamic constraints with both static and dynamic collision constraints. This integration enables simultaneous planning and control, generating an optimal trajectory that balances safety and efficiency~\cite{brito2019model, chen2021interactive, stefanini2024efficient}. Such methods involve two sequential steps at each time-step: first, utilizing a motion model to predict dynamic obstacles in the environment, and second, formulating the robot’s navigation task as an optimal control problem~\cite{chen2021interactive}.
Nevertheless, a critical challenge with these methods lies in guaranteeing safety within the generated trajectories, particularly given their limited capacity to account for uncertainties in perception and motion.

Control Barrier Functions (CBFs)~\cite{jian2023dynamic, lu2024robot} and Chance Constraints (CCs)~\cite{du2011probabilistic, zhu2019chance} are widely recognized as effective methodologies for enhancing safety within MPC-based frameworks by integrating safety-critical constraints into the optimization process.
CBFs dynamically adjust control inputs to maintain system states within predefined safe sets, considering worst-case scenarios of uncertainty to ensure safety and constraint satisfaction; however, they often lead to overly conservative control actions in highly uncertain, dynamic environments, potentially resulting in low system performance or infeasible solutions~\cite{mohamed2023towards}. 
Recent works, including the softening of hard constraints through exact penalties~\cite{lu2024robot}, slack variable relaxations~\cite{zeng2021enhancing}, and generalized CBFs~\cite{ma2021feasibility}, have effectively mitigated CBF infeasibility issues while enhancing safety in crowded and dynamic scenarios.
On the other hand, CCs address the limitations of CBFs by providing probabilistic safety guarantees, allowing limited violations as long as the probability of such violations remains below an acceptable threshold~\cite{mustafa2023probabilistic}.  
Despite their advantages, CCs face challenges such as computational complexity and conservatism associated with probabilistic modeling and fixed thresholds for constraint violations, which constrain their scalability in dynamic and high-dimensional environments~\cite{du2011probabilistic}.
To address these limitations,~\cite{castillo2020real} introduced a computationally efficient real-time framework for chance-constrained optimization by leveraging a simplified probabilistic model to manage dynamic obstacles, while~\cite{zhu2019chance, mustafa2023probabilistic} achieved similar improvements by dynamically adjusting risk thresholds to balance safety and efficiency in real-time. Additionally,~\cite{ryu2024integrating} developed a distributionally robust chance-constrained MPC framework that incorporates Conditional Value at Risk (CVaR) to address predictive motion uncertainties and mitigate the conservatism of Gaussian distribution assumptions.
Yet, as the number of dynamic obstacles increases, the dimensionality of constraints grows rapidly, significantly reducing the performance of MPC and increasing the likelihood of infeasible solutions~\cite{brito2019model, du2011probabilistic}. Furthermore, both system dynamics and probabilistic constraints often require linearization to convert the optimization problem into a convex form, ensuring computational traceability~\cite{zhu2019chance}.

To mitigate the challenges posed by gradient-based MPC methods, this letter introduces a new avenue for sampling-based MPC approaches that make no assumptions or approximations on objective functions and system dynamics, targeting autonomous navigation in crowded environments. 
Specifically, we propose C\textsuperscript{2}U-MPPI, an extension of our U-MPPI algorithm~\cite{mohamed2023towards} which incorporates probabilistic chance constraints into the optimization framework to enhance collision avoidance under uncertainty while preserving computational efficiency.
By leveraging the Unscented Transform (UT) for trajectory sampling and integrating risk-sensitive trajectory evaluation, our method generates more feasible trajectories that account for both static and dynamic obstacles, enabling robust and adaptive navigation in uncertain environments.
The main contributions of this work are summarized as follows:
\begin{enumerate}
    \item A real-time collision avoidance method for dynamic and uncertain environments is proposed by extending the U-MPPI framework with probabilistic collision checking, reformulated in Section~\ref{Robot-Obstacle Collision Avoidance Chance Constraints} into a computationally efficient deterministic form for seamless integration.
    \item 
    Dynamic collision avoidance costs, combining adopted and new formulations, are integrated into MPPI-based algorithms in Section \ref{Cost Function Design}, enabling robust trajectory evaluation and enhancing navigation under dynamic interactions.
    \item A layered dynamic obstacle representation is proposed to efficiently evaluate trajectories over a prediction horizon, enabling real-time handling of multiple obstacles, as detailed in Section \ref{Dynamic Obstacle Representation} and illustrated in Fig.~\ref{fig:predictions-representation}.
    \item The proposed control strategy is validated through simulations and real-world experiments in Section~\ref{Results and Analysis}, demonstrating its effectiveness in both cooperative and non-cooperative scenarios compared to baseline methods. 
\end{enumerate}




\vspace*{-3pt}
\section{Preliminaries}\label{Preliminaries} 
This section presents the core models and tools that support our proposed framework, including robot dynamics, obstacle representations, and trajectory planning formulation.
\vspace*{-2pt}
\subsection{Robot Motion Model}
In this work, we address the problem of trajectory planning and collision avoidance for a single AGV navigating in a 2D unknown environment \(\mathcal{W} \subset \mathbb{R}^2\) with both static and dynamic obstacles. The robot’s dynamics are governed by a nonlinear discrete-time stochastic dynamical system, described as:
\vspace*{-2pt}
\begin{equation}\label{eq: dynamics_system}
   \mathbf{x}_{k+1}=f\left(\mathbf{x}_{k},\mathbf{u}_{k}+\delta \mathbf{u}_{k}\right),   
\end{equation}
where \(\mathbf{x}_k = [\mathbf{p}_k^\top, \theta_k]^\top \in \mathbb{R}^{n_x}\) denotes the robot’s state at time-step \(k\), with position \(\mathbf{p}_k = [x_k, y_k]^\top\) and orientation $\theta_k$, and \(\mathbf{u}_k \in \mathbb{R}^{n_u}\) is the control input.
The control input is perturbed by additive zero-mean Gaussian noise \(\delta \mathbf{u}_k\) with covariance \(\Sigma_{\mathbf{u}}\), modeling uncertainty in the system’s actuation.
Considering a finite time-horizon \(N\), the sequence of control inputs \(\mathbf{U}\) is represented as 
$ \mathbf{U} = \left[\mathbf{u}_{0}, \mathbf{u}_{1}, \dots, \mathbf{u}_{N-1}\right]^{\top} \!\! \in \!\!\mathbb{R}^{n_u N},$ 
while the corresponding state trajectory of the system is 
$ \mathbf{x} = \left[\mathbf{x}_{0}, \mathbf{x}_{1}, \dots, \mathbf{x}_{N}\right]^{\top} \in \mathbb{R}^{n_x (N+1)}.
$
The area occupied by the robot at state \( \mathbf{x}_{k} \), denoted \( \mathcal{X}^{\text{rob}}(\mathbf{x}_{k}) \subset \mathcal{W} \), is approximated as a circular region \( \mathcal{A}(\mathbf{p}_k, r_r) \), centered at \( \mathbf{p}_k \) with radius \( r_r \).
\subsection{Dynamic and Static Obstacles Model}
The environment \(\mathcal{W} \subset \mathbb{R}^2\) contains dynamic obstacles, specifically human pedestrians, modeled as non-cooperative agents with uncertain trajectories. Each dynamic obstacle at time-step \(k\) has a state \(\mathbf{o}_k^n\), where \(n \in \{1, \dots, N_\mathbf{o}\}\), and occupies a circular area \(\mathcal{O}^{\text{dyn}}_{k,n} \subset \mathcal{W}\) with center \(\mathbf{p}_k^n = [x_k^n, y_k^n]^\top\) and radius \(r_o^n\). Obstacle motion follows the transition \(\mathbf{o}_{k+1}^n = \xi(\mathbf{o}_k^n)\). As in \cite{brito2019model}, pedestrian dynamics follow a constant velocity model, where \(\ddot{\mathbf{p}}^n = \omega^n\), and \(\omega^n \sim \mathcal{N}(0, \mathbf{\Sigma}_k^{\mathbf{o}})\) is zero-mean Gaussian noise with diagonal covariance \(\mathbf{\Sigma}^{\mathbf{o}} \in \mathbb{R}^2\), implying that acceleration arises solely from uncertainty.
Using observed position data, the future positions and associated uncertainties of the obstacles are estimated and predicted through a standard Linear Kalman Filter (LKF).
To reflect realistic conditions, we assume the robot operates in an unknown or partially observed environment. Accordingly, a robot-centered 2D occupancy grid (local costmap) is continuously updated using sensory data to represent the surrounding static obstacles, denoted as \(\mathcal{O}^{\text{stc}} \subset \mathcal{W}\).
To prevent overlap between static and dynamic collision constraints and to reduce noise caused by increasing pedestrian numbers in the costmap \cite{mohamed2022autonomous}, (i) detected pedestrians are filtered from the point cloud and classified as dynamic agents, and (ii) the costmap is then generated using only static obstacles.
\subsection{Problem Formulation}
Given the cost function \(J\), robot footprint \(\mathcal{X}^{\text{rob}}(\mathbf{x}_k)\), dynamic obstacles \(\mathcal{O}_k^{\text{dyn}} = \bigcup_{n=1}^{N_\mathbf{o}} \mathcal{O}_{k,n}^{\text{dyn}}\), static obstacles \(\mathcal{O}^{\text{stc}}\), and initial and goal states \(\mathbf{x}_s\) and \(\mathbf{x}_f\), the objective is to compute an optimal control sequence \(\mathbf{U} = \left\{\mathbf{u}_k\right\}_{k=0}^{N-1}\) that safely and efficiently drives the robot from \(\mathbf{x}_s\) to \(\mathbf{x}_f\), while avoiding collisions with both static and dynamic obstacles. The resulting optimization problem is formalized as follows:
\setlength{\jot}{-0pt} 
\begin{subequations}
\vspace{-1pt}
\label{eq:2}
\begin{align}
\min_{\mathbf{U}} \quad J &= \mathbb{E}\left[\phi\left(\mathbf{x}_N\right) + \sum_{k=0}^{N-1} \left(q\left(\mathbf{x}_k\right) + \frac{1}{2} \mathbf{u}_k^\top R \mathbf{u}_k\right)\right]\!, \label{eq:2a} \\
\text{s.t.} \quad & \mathbf{x}_{k+1} = f\left(\mathbf{x}_k, \mathbf{u}_k + \delta \mathbf{u}_k\right), \delta \mathbf{u}_{k} \sim \mathcal{N}(\mathbf{0}, \Sigma_{\mathbf{u}}), \label{eq:2b} \\
&  \mathbf{x}_0 = \mathbf{x}_s, \quad \mathbf{u}_k \in \mathbb{U}, \quad \mathbf{x}_k \in \mathbb{X}, \label{eq:2c} \\
& \mathcal{X}^{\text{rob}}\left(\mathbf{x}_k\right) \cap \big(\mathcal{O}^{\text{stc}} \cup \mathcal{O}_k^{\text{dyn}}\big) = \emptyset, \label{eq:2d}
\end{align}
\vspace{-1pt}
\end{subequations}
where \( J \) denotes the expected value of the trajectory cost, comprising the terminal state cost \( \phi(\mathbf{x}_N) \), an arbitrary state-dependent cost \( q(\mathbf{x}_k) \), and the quadratic control cost \( \frac{1}{2} \mathbf{u}_k^\top R \mathbf{u}_k \), where \( R \in \mathbb{R}^{n_u \times n_u} \) is a positive-definite weighting matrix. 
The optimization accounts for the system dynamics specified in (\ref{eq:2b}), while ensuring that the control inputs \( \mathbf{u}_k \) and states \( \mathbf{x}_k \) remain within their respective feasible sets or constraint sets \( \mathbb{U} \) and \( \mathbb{X} \), as described in (\ref{eq:2c}), and ensures collision avoidance, as outlined in (\ref{eq:2d}).
Recall that the robot and obstacles are modeled as circles, centered at \(\mathbf{p}_k\) and \(\mathbf{p}_k^n\) with radii \(r_r\) and \(r_o^n\), respectively.
Consequently, the minimum safe distance between the robot and obstacle \(\mathbf{o}^n\) is given by \(r_{\text{safe}}^n = r_r + r_o^n\).
Thus, the collision avoidance constraint for dynamic obstacles, stated as \(\mathcal{X}^{\text{rob}}\left(\mathbf{x}_k\right) \cap \mathcal{O}_k^{\text{dyn}} = \emptyset\) in (\ref{eq:2d}), can be formulated as:
\begin{equation}\label{eq: collision-condition-mppi}
    g_n\left(\mathbf{x}_k, \mathbf{o}_k^n\right) =  \left\Vert  \mathbf{x}_k - \mathbf{o}_k^n \right\Vert_2 - r_{\text{safe}}^n, 
\end{equation}
where ensuring that \(g_n(\mathbf{x}_k, \mathbf{o}_k^n) \geq 0\) at each time-step \(k\) guarantees that the robot maintains a safe distance from obstacle \(\mathbf{o}^n\), thus preventing collisions.
In this work, we further extend the problem formulation given by (\ref{eq:2}) by incorporating chance constraints to explicitly account for uncertainty in dynamic obstacles, as detailed in Section~\ref{Robot-Obstacle Collision Avoidance Chance Constraints}.
\vspace*{2pt} 
\section{Overview of U-MPPI Control Strategy}
The vanilla MPPI algorithm \cite{williams2017model} minimizes the cost function \(J\) in (\ref{eq:2}) by sampling \(M\) trajectories, each generated by applying random control perturbations \(\delta \mathbf{u}_{k,m}\) to a nominal control sequence \(\mathbf{U} = \{\mathbf{u}_k\}_{k=0}^{N-1}\). Each trajectory \(\tau_m\), where \(m \in \{1, \dots, M\}\), is forward-simulated over the prediction horizon \(N\) according to the system dynamics and assigned a \textit{cost-to-go} \(S_m\), which quantifies its performance. MPPI then iteratively updates each optimal control input \(\mathbf{u}_k\) using a weighted sum of these perturbations, prioritizing lower-cost trajectories. 
Although MPPI effectively explores the state-space and handles nonlinear dynamics, it assumes uncertainty solely through control perturbations and does not explicitly propagate it through the system states.
To address this limitation, we extend MPPI with the UT, resulting in the Unscented MPPI (U-MPPI) algorithm, as introduced in our previous work \cite{mohamed2023towards}. 
By incorporating the UT, U-MPPI propagates both the mean \(\hat{\mathbf{x}}_k\) and covariance \(\mathbf{\Sigma}_k^{\mathbf{x}}\) of the system dynamics, with \(\mathbf{x}_k \sim \mathcal{N}(\hat{\mathbf{x}}_k, \mathbf{\Sigma}_k^{\mathbf{x}})\), enabling more efficient trajectory sampling and improved state-space exploration.
Additionally, U-MPPI incorporates a risk-sensitive cost function, \(q_{\text{rs}}\), which adjusts control actions based on a risk sensitivity parameter \(\gamma\), where \(\gamma > 0\) promotes \textit{risk-seeking} behavior, \(\gamma < 0\) induces \textit{risk-averse} behavior, and \(\gamma = 0\) recovers the \textit{risk-neutral} case.

Building on this foundation, the U-MPPI control loop operates by estimating the current system state \(\mathbf{x}_{0} \sim \mathcal{N}(\hat{\mathbf{x}}_0, \mathbf{\Sigma}_0^\mathbf{x})\) and generating \(N \times M_\sigma\) random control perturbations \(\delta \mathbf{u}\). It then samples \(M_\sigma\) batches in parallel on the GPU, with each batch containing \(n_\sigma\) trajectories, resulting in a total of \(M = n_\sigma M_\sigma\) rollouts, where \(n_\sigma = 2n_x + 1\) represents the number of sigma points.
For each batch \(m \in \{1, \ldots, M_\sigma\}\), the state distribution \(\mathbf{x}_{k} \sim \mathcal{N}(\hat{\mathbf{x}}_k, \mathbf{\Sigma}_k^{\mathbf{x}})\) is approximated using \(n_\sigma\) sigma points \(\{\mathcal{X}_{k}^{(i)}\}_{i=0}^{2n_x}\) at each time-step \(k\). These sigma points are computed as \(\mathcal{X}_k^{(0)} = \hat{\mathbf{x}}_k\), \(\mathcal{X}_k^{(i)} = \hat{\mathbf{x}}_k + \big(\sqrt{(n_x + \lambda_\sigma)\mathbf{\Sigma}_k^{\mathbf{x}}}\big)_{i}\) for \(i = 1, \ldots, n_x\), and \(\mathcal{X}_k^{(i)} = \hat{\mathbf{x}}_k - \big(\sqrt{(n_x + \lambda_\sigma)\mathbf{\Sigma}_k^{\mathbf{x}}}\big)_{i}\) for \(i = n_x + 1, \ldots, 2n_x\), where \(\lambda_\sigma = \alpha^2(n_x + k_\sigma) - n_x\), with \(\lambda_\sigma\) influenced by the scaling parameters \(k_\sigma \geq 0\) and \(\alpha \in (0,1]\), which control the dispersion of the sigma points around the mean. 
The sigma points are propagated through the underlying nonlinear dynamics described in (\ref{eq: dynamics_system}), as \(\mathcal{X}_{k+1}^{(i)} = f(\mathcal{X}_k^{(i)}, \mathbf{u}_{k} + \delta \mathbf{u}_{k})\), which can be compactly written as \(\mathbf{X}_{k+1} = \mathbf{f}(\mathbf{X}_{k}, \mathbf{u}_{k} + \delta \mathbf{u}_{k})\), where \(\boldsymbol{X}_{k+1} =  \left[\mathcal{X}_{k+1}^{(0)}, \ldots, \mathcal{X}_{k+1}^{(2n_x)}\!\right]^{\top}  \in \mathbb{R}^{n_x n_\sigma}\).
After propagation, the sigma points are converted back into the mean and covariance using \(\hat{\mathbf{x}}_{k+1} = \sum_{i=0}^{2n_x} w_{m}^{(i)}\mathcal{X}_{k+1}^{(i)}\) and \(\mathbf{\Sigma}_{k+1}^{\mathbf{x}} = \sum_{i=0}^{2n_x} w_{c}^{(i)}(\mathcal{X}_{k+1}^{(i)} - \hat{\mathbf{x}}_{k+1})(\mathcal{X}_{k+1}^{(i)} - \hat{\mathbf{x}}_{k+1})^{\top}\), where \(w_m^{(i)}\) and \(w_c^{(i)}\) are weights for computing the mean and covariance, respectively; these weights are defined as \(w_m^{(0)} = \frac{\lambda_\sigma}{n_x + \lambda_\sigma}\), \(w_c^{(0)} = w_m^{(0)} + (1 - \alpha^2 + \beta)\), and \(w_m^{(i)} = w_c^{(i)} = \frac{1}{2(n_x + \lambda_\sigma)}\) for \(i = 1, \ldots, 2n_x\), where \(\beta\) is a hyper-parameter balancing mean and covariance.
This sequential process, repeated over the entire time-horizon \(N\), ensures accurate propagation of both the mean and covariance of the state distribution, yielding a sequence of state vectors \(\{\mathbf{X}_1, \mathbf{X}_2, \ldots, \mathbf{X}_N\} \equiv \big\{\mathbf{X}_k\big\}_{k=1}^{N}\), which represents a batch of \(n_\sigma\) sampled trajectories.

By propagating the system uncertainty \(\mathbf{\Sigma}_{k}^{\mathbf{x}}\), U-MPPI integrates this uncertainty into trajectory evaluation using a risk-sensitive state-dependent cost function \( q_{\mathrm{rs}} \), which replaces the commonly used quadratic cost in MPPI with an uncertainty-aware formulation, as reflected in \( q_{\text{goal}} \) in \eqref{eq:traj-cost}, ensuring a balance between risk and performance.
The explicit form of \( q_{\text{rs}} \) used to assess the \(i\)-th trajectory in each batch is given by:
\begin{equation}\label{Qrs-modified}
q_{\mathrm{rs}}\!\left(\!\mathcal{X}_{k}^{(i)}\!\!, \mathbf{\Sigma}_{k}^{\mathbf{x}}\!\right) =
\frac{1}{\gamma} \log \operatorname{det}\left(\mathbf{I}\!\,+\!\,\gamma Q \mathbf{\Sigma}_{k}^{\mathbf{x}}\right)
+\left\|\mathcal{X}_{k}^{(i)}\!\!-\mathbf{x}_{f}\right\|_{Q_{\mathrm{rs}}}^2\!\!\!\!,
\vspace*{-2pt}
\end{equation}
where \(Q_{\mathrm{rs}}(\mathbf{\Sigma}_{k}^{\mathbf{x}}) = \left(Q^{-1}+\gamma \mathbf{\Sigma}_{k}^{\mathbf{x}}\right)^{-1} \in \mathbb{R}^{n_x \times n_x}\) 
serves as the risk-sensitive penalty matrix, with the risk-sensitivity parameter \(\gamma\) modulating U-MPPI's response to uncertainty.
For instance, when \(\gamma < 0\), the penalty matrix \(Q_{\mathrm{rs}}\) increases with higher system uncertainty \(\mathbf{\Sigma}_{k}^{\mathbf{x}}\), resulting in a larger \(q_{\mathrm{rs}}\) and imposing greater penalties for deviations from the desired state \(\mathbf{x}_{f}\), whereas for \(\gamma = 0\), \( Q_{\mathrm{rs}} \) remains constant and equal to the weighting matrix \( Q \), regardless of the level of uncertainty, reducing \( q_{\mathrm{rs}} \) to the standard quadratic cost, representing a \textit{risk-neutral} treatment of uncertainty during optimization.
The \textit{cost-to-go} of trajectory \(\tau^{(i)}_m\) in batch \(m\) is defined as:
\setlength{\jot}{1pt} 
\begin{equation}\label{eq:cost-to-go-umppi}
\begin{aligned}
 S\left(\tau^{(i)}_m \right) = &\,\phi\left(\mathcal{X}_{N}^{(i)} \right) + \sum_{k=0}^{N-1} \tilde{q}\left(\mathcal{X}_{k}^{(i)}, \mathbf{\Sigma}_{k}^{\mathbf{x}}, \mathbf{u}_{k}, \delta \mathbf{u}_{k,m}\right), \\ 
 \end{aligned}
\end{equation}
where \(\tilde{q}\) consists of the state-dependent cost \( q \), which is now formulated as a function of the risk-sensitive cost \( q_{\mathrm{rs}} \), as detailed in Section \ref{Cost Function Design}, along with the quadratic control cost \( q_{\mathbf{u}}(\mathbf{u}_{k}, \delta \mathbf{u}_{k,m}) \), which regulates control effort.
In this work, we adopt the control cost $q_{\mathbf{u}}$ from \cite[Eq.~(4)]{mohamed2023towards}, originally proposed in \cite{williams2017model}.
The control input sequence in U-MPPI is updated using the same weighted averaging method as in MPPI:
\begin{equation}
 \mathbf{u}_{k} \!\leftarrow \! \!\mathbf{u}_{k}
        \!+\!
        \frac{\sum_{m=1}^{M_\sigma} \exp \Bigl( \!\frac{-1}{\lambda} \!\bigl(\mathbf{S}\left(\Tau_{m}\right) -S_{\min} \bigr) \Bigr) \delta \mathbf{u}_{k, m}}{\sum_{m=1}^{M_\sigma} \exp \Bigl(\frac{-1}{\lambda} \bigl(\mathbf{S}\left(\Tau_{m}\right) -S_{\min} \bigr)\Bigr)},
        \vspace*{-3pt}
\end{equation}
where \(\mathbf{S}(\Tau_{m}) = \left[S(\tau^{(0)}_{m}), \ldots, S(\tau^{(2n_x)}_{m})\right]^\top \in \mathbb{R}^{n_\sigma}\) represents the costs of all sigma-point trajectories within batch \(m\).
The optimal control sequence \(\{\mathbf{u}_{k}\}_{k=0}^{N-1}\) is smoothed with a Savitzky-Golay filter, then the first control \(\mathbf{u}_0\) is applied to the system, and the remaining \(N-1\) steps are shifted as a warm-start for the next control loop.
For further details on U-MPPI, including the handling of task-related state, control, and collision avoidance constraints, as well as the impact of parameters \(\gamma\), \(\alpha\), and \(k_\sigma\) on its performance, refer to \cite{mohamed2023towards}.


\vspace*{-1pt}
\section{Proposed C\textsuperscript{2}U-MPPI Framework}
\vspace*{-1pt}
In this section, we present our proposed chance-constrained U-MPPI control strategy, incorporating dynamic costs and a layered obstacle representation for navigation under uncertainty.
\subsection{Chance-Constrained U-MPPI Problem Formulation}

MPPI enforces the collision avoidance condition in \eqref{eq: collision-condition-mppi}, which assumes deterministic obstacle and robot positions based on known or predicted locations, ensuring collision-free sampled trajectories but disregarding uncertainties in position estimates. In contrast, our U-MPPI framework \cite{mohamed2023towards} incorporates the UT to model uncertainty, making it more suitable for uncertain environments; however, it accounts only for static obstacles, omitting environmental uncertainty and lacking a probabilistic framework for assessing collision risks and managing dynamic obstacle avoidance.
To overcome these limitations, we introduce C\textsuperscript{2}U-MPPI, which embeds Chance Constraints (CCs) \cite{du2011probabilistic} into U-MPPI, replacing the deterministic collision condition with probabilistic collision checking. This is formulated as a chance constraint \(\Pr(C) \leq \delta\) in \eqref{eq:uoc_c}, where \(C\left(\mathbf{p}_k, \mathbf{p}^n_k\right): \|\mathbf{p}_k - \mathbf{p}^n_k\| \leq r_{\text{safe}}^n\) defines the collision condition between the robot position \(\mathbf{p}_k\) and the \(n\)-th obstacle position \(\mathbf{p}^n_k\), with \(\delta \in \mathbb{R}^{+}\) specifying the confidence level for avoidance.
Accordingly, the stochastic optimal control problem in (\ref{eq:2}) is reformulated within the C\textsuperscript{2}U-MPPI framework as follows:
\begin{subequations}
\begin{align}
\min _{\mathbf{U}} \; \mathbf{J} (\mathbf{X},\mathbf{u}) &=  \mathbb{E}\!\left[\mathbf{\Phi 
 }\!\left(\mathbf{X}_{N}\!\right)\!+\!\!\!\sum_{k=0}^{N-1}\!\!\left(\!\!\mathbf{q}\!\left(\mathbf{X}_{k}\right)\!+\!\frac{1}{2} \mathbf{u}_{k}^{\top}  R \mathbf{u}_{k}\!\!\right)\!\!\right]\!\!, \label{eq:uoc_a}\\
\text {s.t.} 
\quad \mathbf{X}_{k+1} &=\mathbf{f}\left(\mathbf{X}_{k}, \mathbf{u}_{k} + \delta \mathbf{u}_{k}\right), \delta \mathbf{u}_{k} \sim \mathcal{N}(\mathbf{0}, \Sigma_{\mathbf{u}}), \label{eq:uoc_b}\\
 \Pr(C&) \leq \delta \quad \text{and} \quad \mathcal{X}^{\text{rob}}(\mathbf{X}_k) \cap \mathcal{O}^{\text{stc}} = \emptyset,
\label{eq:uoc_c}\\ 
\mathbf{X}_0 = &\left[\mathcal{X}_0^{(0)}\!, \ldots, \mathcal{X}_0^{(2n_x)}\right]^{\top} \!\!, \mathbf{u}_{k} \!\in \mathbb{U},\mathbf{X}_{k} \!\in \mathbb{X},
\end{align}
\label{eq: unscented optimal control problem}
\end{subequations} 
where $\mathbf{\Phi} \left(\mathbf{X}_{N} \right)\!\! = \!\!\left[\phi \left(\mathcal{X}_{N}^{(0)}\right), \ldots, \phi \left(\mathcal{X}_{N}^{(2n_x)}\right)\right]^{\top}  \!\!\! \in \! \mathbb{R}^{n_\sigma}$ 
and 
$\mathbf{q} \left(\mathbf{X}_k \right) = \left[q\left(\mathcal{X}^{(0)}_k \right), \ldots, q\left(\mathcal{X}^{(2n_x)}_k \right)\right]^{\top}  \!\! \in \mathbb{R}^{n_\sigma}$.

\subsection{Robot-Obstacle Collision Avoidance Chance Constraints}\label{Robot-Obstacle Collision Avoidance Chance Constraints}

Solving \eqref{eq: unscented optimal control problem} with the chance constraint \(\Pr(C) \leq \delta\) is computationally prohibitive, as it requires evaluating collision probabilities for all sampled trajectories under uncertainty, where the joint distributions are complex and intractable for real-time feasibility. To address this, we aim to reformulate this constraint as a deterministic condition on the robot’s and obstacle’s mean states. To this end, we rely on the probabilistic framework from \cite{du2011probabilistic} to approximate \(\Pr(C)\), ensuring computational efficiency in the C\textsuperscript{2}U-MPPI optimization process.
Following \cite{du2011probabilistic}, we simplify the analysis by modeling the robot as a disk with area \( \mathcal{A}(\mathbf{p}_k, r_r) \) and the obstacle as a point at \( \mathbf{p}_k^n = [x_k^n, y_k^n]^\top \).
Accordingly, the collision condition is given by \( C: \mathbf{p}^n_k \in \mathcal{A}(\mathbf{p}_k, r_r) \).
The instantaneous collision probability is derived from the joint distribution of the robot and obstacle positions:
\begin{equation}
\Pr(C) = \int_{\mathbf{p}_k} \int_{\mathbf{p}^n_k} \mathbb{I}_c(\mathbf{p}^n_k, \mathbf{p}_k) \, p(\mathbf{p}_k, \mathbf{p}^n_k) \, d\mathbf{p}_k \, d\mathbf{p}^n_k,
\end{equation}
where \(p(\mathbf{p}_k, \mathbf{p}^n_k) = p(\mathbf{p}^n_k \mid \mathbf{p}_k) \, p(\mathbf{p}_k)\) represents the joint likelihood of the robot being at \(\mathbf{p}_k\) and the obstacle at \(\mathbf{p}^n_k\) and \(\mathbb{I}_c\) is the indicator function defined as:
\begin{equation}
\mathbb{I}_c(\mathbf{p}^n_k, \mathbf{p}_k) = \begin{cases}
1, & \text{if } \mathbf{p}^n_k \in \mathcal{A}(\mathbf{p}_k, r_r), \\
0, & \text{otherwise.}
\end{cases}
\end{equation}
By applying the indicator function \(\mathbb{I}_c\) and expanding the joint distribution, the probability of collision can be formulated as:
\begin{equation}\label{eq:applyI_c}
\Pr(C)\! = \!\int_{\mathbf{p}_k} \!\!\Big[\int_{\mathbf{p}^n_k \in \mathcal{A}(\mathbf{p}_k, r_r)} \!\!p(\mathbf{p}^n_k \mid \mathbf{p}_k) \, d\mathbf{p}^n_k \Big] p(\mathbf{p}_k) \, d\mathbf{p}_k.
\end{equation}

To efficiently solve the integral in \eqref{eq:applyI_c}, we introduce two simplifying assumptions. First, reflecting the inherent compact size of the robot, we approximate the inner integral in \eqref{eq:applyI_c} by evaluating the conditional distribution \(p(\mathbf{p}_k^n = \mathbf{p}_k \mid \mathbf{p}_k)\) of the obstacle at the robot’s position \(\mathbf{p}_k\), multiplied by the area \(A_r = \pi r_r^2\) of the region \(\mathcal{A}(\mathbf{p}_k, r_r)\). Second, consistent with the Gaussian modeling approach adopted in the C\textsuperscript{2}U-MPPI framework, we assume the positional uncertainties of the robot and obstacle are independent Gaussian distributions: \(\mathbf{p}_k \sim \mathcal{N}(\hat{\mathbf{p}}_k, \mathbf{\Sigma}_k^{\mathbf{x}})\) and \(\mathbf{p}_k^n \sim \mathcal{N}(\hat{\mathbf{p}}_k^n, \mathbf{\Sigma}_k^{\mathbf{o}})\). Under these assumptions, and following the derivations in \cite{du2011probabilistic},
the probability of collision can be approximated as:
\begin{equation}\label{eq:P_C-approximation}
\Pr(C) \approx \frac{A_r}{\eta_c} \exp \Big( -\frac{1}{2} (\hat{\mathbf{p}}_k - \hat{\mathbf{p}}^n_k)^{\top} \mathbf{\Sigma}_k^{\mathbf{c}^{-1}} (\hat{\mathbf{p}}_k - \hat{\mathbf{p}}^n_k) \Big),
\end{equation}  
where 
\(\mathbf{\Sigma}_k^{\mathbf{c}} \triangleq \mathbf{\Sigma}_k^{\mathbf{x}} + \mathbf{\Sigma}_k^{\mathbf{o}}\)
denotes the combined covariance of the two distributions, and \(\eta_c \triangleq \sqrt{\det(2 \pi \mathbf{\Sigma}_k^\mathbf{c})}\).
Building on the derived approximation in~\eqref{eq:P_C-approximation}, the chance constraint for collision avoidance, \( \Pr(C) \leq \delta \), can be expressed as:
\begin{equation}\label{eq:CC-condition}
\mathcal{M}(\hat{\mathbf{p}}_k, \hat{\mathbf{p}}^n_k, \mathbf{\Sigma}_k^\mathbf{c}) = \left\|\hat{\mathbf{p}}_k - \hat{\mathbf{p}}^n_k\right\|_{\mathbf{\Sigma}_k^{\mathbf{c}^{-1}}}^2 \geq \kappa\left(\mathbf{\Sigma}_k^\mathbf{c}, A_r, \delta\right),
\end{equation}
where 
\(
\mathcal{M} = \left(\hat{\mathbf{p}}_k - \hat{\mathbf{p}}^n_k\right)^{\top} \mathbf{\Sigma}_k^{\mathbf{c}^{-1}}\left(\hat{\mathbf{p}}_k - \hat{\mathbf{p}}^n_k\right)\) is the squared Mahalanobis distance, and \(\kappa = -2 \ln \left(\frac{\eta_c \,\delta}{ A_r }\right)\) is a deterministic threshold computed at each time-step \(k\) from \(\mathbf{\Sigma}_k^\mathbf{c}\), \(A_r\), and \(\delta\).
The inequality in \eqref{eq:CC-condition} acts as a deterministic collision avoidance constraint, requiring the robot's mean position \(\hat{\mathbf{p}}_k\) to remain outside the ellipsoidal safety boundary centered at the mean position of the obstacle \(\hat{\mathbf{p}}^n_k\). As a result, the probabilistic constraint \( \Pr(C) \leq \delta \) in \eqref{eq:uoc_c} is effectively reformulated into a computationally feasible deterministic form that integrates seamlessly into our proposed control framework, ensuring safe navigation under uncertainty.
\begin{figure}[!t]%
    \vspace*{-5pt}
    \centering
    \subfloat{%
        \includegraphics[scale=0.99]{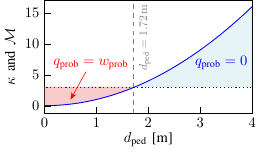}%
        \label{fig:collision_avoidance_identity}}%
    \hspace*{-5pt}
    \subfloat{%
        \includegraphics[scale=0.99]{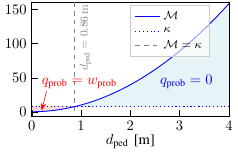}%
        \label{fig:collision_avoidance_tenth}}%
    \vspace*{-2pt}
    \caption{Collision-check behavior vs. \(\!d_\text{ped}\) under two uncertainty levels:~\(\mathbf{\Sigma}^\mathbf{o} \!\!=\! \mathbf{I}_2\) (left) and \(0.1\mathbf{I}_2\) (right), with \(r_r \!\!=\!\! \SI{0.3}{\metre}\) and \(\delta \!\!=\!\! 0.01\). Shaded regions denote constraint satisfaction (blue) and violation (red).}
    \label{fig:collison-codition-plot}%
    \vspace*{-13pt}
\end{figure}

Notably, unlike many approaches that uniformly inflate robot and obstacle geometries with fixed pre-defined inflation to construct conservative safety margins (e.g., \cite{brito2019model}) or rely on linearizing chance constraints to retain convexity (e.g., \cite{mustafa2023probabilistic, zhu2019chance}), our method enables a more precise, direction-aware safety buffer through a nonlinear and nonconvex formulation that dynamically assesses and updates the combined uncertainty \(\mathbf{\Sigma}_k^{\mathbf{c}}\) in real time.

To illustrate the sensitivity of the collision constraint in \eqref{eq:CC-condition} to uncertainty, Fig.~\ref{fig:collison-codition-plot} shows how \(\mathcal{M}\) and \(\kappa\) vary with the Euclidean distance \(d_\text{ped} = \|\hat{\mathbf{p}}_k - \hat{\mathbf{p}}^n_k\|\) under varying pedestrian uncertainty \(\mathbf{\Sigma}^\mathbf{o}\).
As \(\mathbf{\Sigma}^\mathbf{o}\) is reduced from \(\mathbf{I}_2\) (left plot) to \(0.1\mathbf{I}_2\) (right plot), the constraint is satisfied at a significantly shorter distance (\SI{0.86}{\metre} instead of \SI{1.72}{\metre}), thereby reducing the robot’s reaction range and delaying its response to potential collisions. While this reduces conservativeness, it may compromise safety. In contrast, higher uncertainty induces earlier reactions and wider safety margins, albeit at the expense of efficiency, such as longer paths and increased control effort. This trade-off directly impacts trajectory evaluation in C\textsuperscript{2}U-MPPI, where constraint violations (i.e., \(\mathcal{M} < \kappa\)) are penalized through the probabilistic cost \(q_{\text{prob}}\) in \eqref{eq:prop-cost} and correspond to the red-shaded regions.

\subsection{Cost Function Design}\label{Cost Function Design}
In sampling-based MPC approaches, the state-dependent running cost \(q(\cdot)\) serves as a key component in directing the robot's behavior by penalizing undesired states.
For safe, goal-driven navigation in a 2D environment with static and dynamic obstacles, we evaluate each sampled trajectory as:
\begin{equation}\label{eq:traj-cost}
q(\cdot) = q_{\text{goal}}(\cdot) + q_{\text{stc}}(\cdot) + q_{\text{dyn}}(\cdot),
\end{equation}
which is composed of three terms: goal-directed cost \(q_{\text{goal}}(\cdot)\), and collision avoidance costs for static and dynamic obstacles, \(q_{\text{stc}}(\cdot)\) and \(q_{\text{dyn}}(\cdot)\), respectively.
Here, \((\cdot)\) in \(q(\cdot)\) specifically denotes \((\mathbf{x}_k)\) in MPPI, while in C\textsuperscript{2}U-MPPI, including classical U-MPPI, it refers to \((\mathcal{X}_{k}^{(i)}, \mathbf{\Sigma}_{k}^{\mathbf{x}})\).
The first term \(q_{\text{goal}}(\cdot)\) penalizes the robot's deviation from the desired goal state \( \mathbf{x}_f \). 
In MPPI, \(q_{\text{goal}}\) is expressed as the standard quadratic cost, given by \(q_{\text{goal}}\left(\mathbf{x}_k\right) = \left\|\mathbf{x}_k - \mathbf{x}_{f}\right\|_{Q}^2\), whereas the risk-sensitive cost \( q_{\mathrm{rs}}\), as specified in \eqref{Qrs-modified}, is employed in both U-MPPI and its extension C\textsuperscript{2}U-MPPI.
In all control strategies, the second term \(q_{\text{stc}}\) is defined as \(q_{\text{stc}}(\mathbf{x}) = w_{\text{stc}} \mathbb{I}_{\text{stc}}\), acting as an indicator function that applies a high penalty weight \(w_{\text{stc}}\) when the robot collides with static obstacles, where \(\mathbb{I}_{\text{stc}} = 1\) if \(\mathcal{X}^{\text{rob}}(\mathbf{x}) \cap \mathcal{O}^{\text{stc}} \neq \emptyset\), with \(\mathbf{x}\) representing \(\mathbf{x}_k\) in MPPI and \(\mathcal{X}_{k}^{(i)}\) in U-MPPI-based methods. 
In \cite{mohamed2022autonomous}, we employed the first two terms of \(q(\cdot)\) to achieve collision-free navigation in dynamic environments using a 2D local costmap from local perception. 
However, as the number of moving agents increased, the costmap became noisier, increasing the risk of local minima traps.

To address these limitations and enhance safety in dynamic environments, we introduce a dynamic collision avoidance cost, \( q_{\text{dyn}} \), formulated by combining one or more cost functions. In this work, we specifically propose the exponential-shaped cost \( q_{\text{exp}} \) and the probabilistic collision-checking cost \( q_{\text{prob}} \), and incorporate the repulsive cost \( q_{\text{rep}} \) from \cite{brito2019model} to enhance obstacle clearance, thereby offering a more robust and reliable navigation solution.
The respective cost functions are mathematically defined as follows:
\setlength{\jot}{-0pt} 
\begin{align}
    &q_\text{exp}(\mathbf{x}_k) = w_\text{exp} \sum_{n=1}^{N_\mathbf{o}} \exp \biggl( -\alpha_\text{exp} \Bigl(\|\mathbf{p}_k - \mathbf{p}^n_k\| - r_{\text{safe}}^n\Bigr)\!\!\biggr), \label{eq:exp-cost} \\
    &q_{\text{rep}}(\mathbf{x}_k) = w_\text{rep} \sum_{n=1}^{N_\mathbf{o}} \biggl(\frac{1}{\|\mathbf{p}_k - \mathbf{p}^n_k\|^2 + \gamma_\text{rep}}\biggr), \label{eq:rep-cost}\\
    &q_{\text{prob}}(\mathcal{X}_{k}^{(i)}, \mathbf{\Sigma}_k^\mathbf{x}) = w_{\text{prob}} \begin{cases}
1, & \text{if} \ \mathcal{M}< \kappa, \label{eq:prop-cost} \\
0, & \text{otherwise},
\end{cases}
\end{align}
where \(w_{\text{exp}}, 
w_{\text{rep}}\) and \(w_{\text{prob}}\) are the collision penalty weighting coefficients determining the influence of each cost function, \(\alpha_{\text{exp}}\)
adjusts the sharpness of the exponential
penalties, and \(\gamma_{\text{rep}}\) serves as a stabilizing constant to prevent singularities in \(q_{\text{rep}}\).
The first trajectory assessment function, \(q_{\text{exp}}\), applies a heavily weighted penalty to trajectories that violate the collision avoidance condition \(g_n(\mathbf{x}_k, \mathbf{o}_k^n) \geq 0\) in \eqref{eq: collision-condition-mppi}, enforcing strict avoidance through a sharp exponential penalty near obstacles. 
The second, \(q_{\text{rep}}\), employs a potential function with an inverse-square penalty to repel the robot from nearby agents.
Additionally, \(q_{\text{rep}}\) enhances clearance from moving obstacles, providing an extra layer of safety and balancing avoidance with smooth motion planning, enabling effective navigation without abrupt maneuvers \cite{brito2019model}.
In contrast, the third objective, \(q_{\text{prob}}\), imposes a stringent binary penalty based on the probabilistic collision-checking condition in \eqref{eq:CC-condition}, heavily penalizing trajectories 
where the margin \(\mathcal{M}\) falls below the risk threshold \(\kappa\), thereby promoting safety under uncertainty, as highlighted by the red-shaded regions in Fig.~\ref{fig:collison-codition-plot}.
While all three assessment functions are deterministic, the first two (\(q_{\text{exp}}\) and \(q_{\text{rep}}\)) are applicable to all control strategies, where in U-MPPI-based methods, \(\mathbf{x}_k\) and \(\mathbf{p}_k\) are replaced by \(\mathcal{X}_{k}^{(i)}\) and \(\mathbf{p}_k^{(i)}\); whereas the third, \(q_{\text{prob}}\), is exclusive to C\textsuperscript{2}U-MPPI due to its reliance on the propagated robot and obstacle uncertainties, \(\mathbf{\Sigma}_k^{\mathbf{x}}\) and \(\mathbf{\Sigma}_k^\mathbf{o}\).

\subsection{Dynamic Obstacle Representation}\label{Dynamic Obstacle Representation}
In gradient-based MPC algorithms such as MPCC \cite{brito2019model}, dynamic obstacles are addressed by explicitly incorporating their predicted positions as constraints in the trajectory planning optimization process, ensuring collision-free navigation by avoiding intersections with the predicted obstacle paths. However, as the number of obstacles increases, the corresponding rise in constraints significantly expands the problem’s dimensionality, leading to higher computational demands and thus limiting the number of obstacles the algorithm can effectively manage.
To overcome this limitation, we propose a more scalable solution that leverages the parallel nature of sampling and GPU-based computation: a layered dynamic obstacle representation tailored specifically for sampling-based MPC frameworks.
While inspired by the general concept of time-indexed obstacle prediction in spatiotemporal occupancy grid maps (SOGMs)~\cite{thomas2022learning, wu2024decentralized}, our method fundamentally differs in both structure and usage.
It discretizes predicted obstacle states over the control horizon \(N\), employs a lightweight LKF to propagate dynamic obstacles, and constructs a per-step layered representation that supports efficient parallel trajectory evaluation without the need for continuous-time optimization~\cite{wu2024decentralized} or neural-based forecasting techniques~\cite{thomas2022learning}. As illustrated in Fig.~\ref{fig:layered-obstacle}, each layer corresponds to a specific future time step and holds the predicted obstacle positions for that time.
Rather than directly constraining predicted positions from the LKF, as in Fig.~\ref{fig:person-predictions}, we evaluate sampled trajectories based on the corresponding obstacle layer at each time step. For instance, at \(k = 3\), the control algorithm considers only the third layer of predictions to compute \(q_{dyn}(.)\). To address discrepancies between the prediction horizon of the pedestrians and the controller, linear interpolation is used to align them before constructing the layers.
Such a solution enables the incorporation of more obstacles in the optimization process while preserving efficient trajectory evaluation and real-time performance in dynamic and uncertain environments, as demonstrated in the supplementary video featuring 30 dynamic pedestrians.
\vspace*{-2pt}
\section{Results and Analysis}\label{Results and Analysis}
This section evaluates our proposed C\textsuperscript{2}U-MPPI in simulated corridors with varying pedestrian densities and through real-world demonstrations in a human-shared indoor environment.
\subsection{Simulation-Based Evaluation}\label{Simulation Details and Results}
\subsubsection{Simulation Setup:}\label{Simulation Setup:Cluttered Environments}
In this study, extensive simulations are conducted using the fully autonomous ClearPath Jackal robot, the differential-drive kinematic model described in \cite{mohamed2022autonomous}. This kinematic model, with the system state \( \mathbf{x} = [{x}, {y}, \theta]^\top \) and control input \( \mathbf{u} = [v, \omega]^\top \), representing the robot's linear and angular velocities, is used to sample trajectories in MPPI and propagate sigma points in U-MPPI-based methods.
We employ Pedsim\footnote{Pedestrian simulator code: \url{https://github.com/srl-freiburg/pedsim_ros}}, an open-source ROS simulator
based on the social force model, to simulate realistic pedestrian interactions.

To demonstrate the advantages of sampling-based MPC over gradient-based methods, we evaluate C\textsuperscript{2}U-MPPI against U-MPPI~\cite{mohamed2023towards}, log-MPPI~\cite{mohamed2022autonomous}, and MPCC~\cite{brito2019model}.
To ensure a fair comparison, simulations for U-MPPI, log-MPPI, and C\textsuperscript{2}U-MPPI were conducted using the same hyperparameter settings specified in \cite{mohamed2023towards}, with two exceptions: (i) a \SI{7}{\second} prediction horizon and \SI{30}{\hertz} control frequency were used, yielding \(N = 210\); and (ii) for log-MPPI, the hyperparameters \(\lambda\), \(\Sigma_n\), and 
\(R\) 
were set based on extensive simulations in~\cite{mohamed2022autonomous} to improve performance.
For MPCC, the configuration from~\cite{brito2019model} was re-tuned for improved performance under our test conditions.
All MPPI variants were executed in \textit{real-time} on an NVIDIA GeForce GTX 1660 Ti laptop GPU using the PyCUDA module and integrated with the ROS framework. 
The cost function parameters are as follows: \(q_{\text{goal}}\) uses \(Q = \operatorname{Diag}(8,8,2)\); \(q_{\text{stc}}\) is weighted by \(w_{\text{stc}} = 10^3\); \(q_{\text{exp}}\) uses \(w_{\text{exp}} = 500\), \(\alpha_{\text{exp}} = 40\), and \(r^n_{\text{safe}} = \SI{1}{\metre}\) in the lower-speed scenario (i.e., \textit{Scenario~\#1} in Section~\ref{Simulation Scenarios and Performance Metrics}), and \(\SI{1.1}{\metre}\) for high-speed scenarios (\textit{Scenarios~\#2 and~\#3}); \(q_{\text{rep}}\) uses \(w_{\text{rep}} = 50\), \(\gamma_{\text{rep}} = 0.3\); and \(q_{\text{prob}}\) is configured with \(w_{\text{prob}} = 10^3\), \(\delta = 0.01\), \(r_r = \SI{0.2}{\metre}\), \(\mathbf{\Sigma}_0^\mathbf{x} = 0.001\mathbf{I}_3\), and \(\mathbf{\Sigma}_0^\mathbf{o} = \mathbf{I}_2\), where \(\mathbf{I}_n\) is the \(n \times n\) identity matrix.

\subsubsection{Simulation Scenarios and Performance Metrics: }\label{Simulation Scenarios and Performance Metrics} 

For performance evaluation, we utilize a \(\SI{30}{\metre} \times \SI{6}{\metre}\) corridor environment with either 6 or 10 pedestrians moving in opposite directions at varying speeds to simulate realistic interactions (see the supplementary material section at the end of this manuscript for details). 
To evaluate the proposed strategies' performance, we consider three scenarios with varying reference speeds for both the robot and pedestrians, along with two distinct interaction modes.  
In the first scenario, denoted as \textit{Scenario \#1} \((30\%, 1)\), the robot’s maximum speed \(v_{\max}\) is set to \SI{1.0}{\metre\per\second}, with each pedestrian moving at 30\% of their maximum speed. In \textit{Scenario \#2} \((80\%, 1.5)\), \(v_{\max}\) is increased to \SI{1.5}{\metre\per\second}, while pedestrians move at 80\% of their maximum speed; in \textit{Scenario \#3} \((100\%, 2)\), \(v_{\max}\) is set to \SI{2.0}{\metre\per\second}, while pedestrians operate at full speed.
Additionally, each scenario includes two interaction modes: \textit{non-cooperative mode (NCM)}, where pedestrians move independently without regard for the robot, and \textit{cooperative mode (CM)}, where pedestrians adapt their trajectories to avoid the robot using social force dynamics.
To ensure a fair comparison, two key aspects are considered across all simulations: (i) the robot is tasked with sequentially reaching predefined target poses, starting at \(G_1 = [0 \, \si{\meter}, -1 \, \si{\meter}, 0^\circ]\), then \(G_2 = [30 \, \si{\meter}, 1 \, \si{\meter}, 100^\circ]\), and finally returning to the origin, thereby promoting dynamic interactions with pedestrians to evaluate the robustness of the control strategies; and (ii) the dynamic collision cost is defined as \(q_{\text{dyn}} = q_{\text{exp}} + q_{\text{rep}}\) in both log-MPPI and U-MPPI, with \(\gamma = 1\) (denoted as U-MPPI\textsuperscript{1}) and \(\gamma = -1\) (U-MPPI\textsuperscript{-1}), 
while C\textsuperscript{2}U-MPPI applies \(q_{\text{dyn}} = q_{\text{prob}}\) with a 2-second pedestrian prediction horizon, denoted as $N_p = \SI{2}{\second}$. 
It is essential to highlight that while the dynamic obstacle representation from Section~\ref{Dynamic Obstacle Representation} is compatible with log-MPPI and the two U-MPPI variants, our evaluation excluded any pedestrian prediction horizon, thus demonstrating robustness relative to MPCC, which applies \( N_p = \SI{3}{\second}\). 

To assess autonomous navigation in dynamic settings, we employ metrics for motion safety, trajectory quality, and control smoothness, focusing exclusively on the robot’s trajectory and excluding pedestrian comfort, as we prioritize NCM over CM (see Section~\ref{Simulation-Results}). Motion safety, represented by the number of collisions \( \mathcal{N}_{\text{c}} \) 
 and velocity constraint violation rate \( \mathcal{V}_{\text{viol}} \), indicates obstacle avoidance effectiveness and adherence to safe operating speeds, while robot trajectory quality, measured by the average distance traversed \( d_{\text{av}} \), average robot speed \( v_{\text{av}} \), and average dynamic energy \( E_{\text{dyn}} \), captures path efficiency and energy use. Control smoothness, characterized by average accumulated jerk \( J_{\text{acc}} \) and average execution time per iteration \( t_{\text{exec}} \), reflects the control system's stability and real-time adaptability.
As most metrics are detailed in \cite{mohamed2023towards}, only \( E_{\text{dyn}} \), \( J_{\text{acc}} \), and \( \mathcal{V}_{\text{viol}} \) are defined here. 
\( E_{\text{dyn}} \!= \!\frac{1}{N_t} \sum_{i=1}^{N_t} \!\big( \frac{v_{\text{peak}} - v_i}{v_{\text{peak}}} \big)^2 \) quantifies energy consumption by averaging the deviation of instantaneous speed \( v_i \) from the peak speed \( v_{\text{peak}} = \max_{i \leq N_t} v_i \), while \( J_{\text{acc}}\! = \!\frac{1}{N_t} \!\sum_{i=1}^{N_t} j_i \), 
where \( N_t \) denotes the number of discrete time steps in the trajectory, and \( j_i \) represents the instantaneous jerk.
\( \mathcal{V}_{\text{viol}} \!=\! \frac{N_{\text{viol}}}{N_t} \times 100\% \) measures the percentage of time steps \( N_{\text{viol}} \) at which the velocity constraint is violated, i.e., \( |v_i|> v_{\text{max}} \).
Lower values of \( E_{\text{dyn}} \), \( J_{\text{acc}} \), and \( \mathcal{V}_{\text{viol}} \) indicate efficient energy use, smooth control, and compliance with the velocity constraint.

\subsubsection{Simulation Results:}\label{Simulation-Results}
\ihab{
Tables \ref{6-ped-NCM} and \ref{10-ped-NCM} present 10-trial averaged performance statistics for the proposed control strategies in goal-oriented navigation, evaluated across our predefined scenarios within 6- and 10-pedestrian environments under NCM. Note that CM is excluded from this comparative analysis due to its inadequacy in providing a rigorous robustness evaluation, as detailed in the supplementary material.
Accordingly, Table~\ref{6-ped-NCM} reports the 6-pedestrian NCM results, comparing C\textsuperscript{2}U-MPPI with log-MPPI and MPCC.
}
 In this uncooperative setting, our C\textsuperscript{2}U-MPPI consistently outperforms both baselines in collision avoidance, achieving zero collisions across all scenarios. It also demonstrates lower \(J_{\text{acc}}\) and \(d_{\text{av}}\), particularly compared to MPCC, while maintaining a higher speed profile in \textit{Scenarios \#2} and \#3, reflecting smoother control actions and more efficient path planning at higher speeds.
While log-MPPI achieves slightly lower \( E_{\text{dyn}} \) in the first two scenarios, C\textsuperscript{2}U-MPPI effectively balances collision avoidance with safe and efficient robot trajectory in complex, uncooperative environments, thanks to its integration of the pedestrian prediction horizon and uncertainty propagation within the dynamic cost \(q_\text{dyn}\).
\begin{table}[t!]
\caption{Performance comparison of C\textsuperscript{2}U-MPPI and two baseline methods (log-MPPI and MPCC) in 6-pedestrian NCM.}
\footnotesize\addtolength{\tabcolsep}{-5.2pt} 
\setlength\extrarowheight{1pt}
\centering
\begin{tabular}{|c|c|c|c|c|c|c|}
\hline
Scheme & $\mathcal{N}_{\text{c}}$ & $E_{\text{dyn}}$ & \(10 \!\times\! J_{\text{acc}}\!\) [\si{\newton}] & $v_{\text{av}}$ [\si{\meter\per\second}] & $d_{\text{av}}$ [\si{\meter}] & $t_{\text{exec}}$ [\si{\milli\second}] \\
\hline\hline
\multicolumn{7}{|c|}{\textit{\textbf{Scenario \#1:}} \((30\%, 1)\), Non-Cooperative Mode (NCM)} \\
\hline
MPCC & 1 & $0.64 \pm 0.17$ & $38.0 \pm 9.9$ & \cellcolor{lightgray}$0.87 \pm 0.06$ & $67.3 \pm 1.6$ & $4.8 \pm 0.8$ \\
log-MPPI & \cellcolor{lightgray}0 & \cellcolor{lightgray}$0.24 \pm 0.03$ & $12.0 \pm 3.9$ & $0.86 \pm 0.01$ & $63.0 \pm 0.9$ & $8.0 \pm 0.4$ \\
Ours & \cellcolor{lightgray}0 & $0.27 \pm 0.04$ & \cellcolor{lightgray}$10.2 \pm 1.4$ & $0.86 \pm 0.01$ & \cellcolor{lightgray}$62.9 \pm 1.1$ & $9.0 \pm 0.2$ \\
\hline\hline
\multicolumn{7}{|c|}{\textit{\textbf{Scenario \#2:}} \((80\%, 1.5)\), Non-Cooperative Mode (NCM)} \\
\hline
MPCC & 7 & $0.70 \pm 0.20$ & $50.2 \pm 9.7$ & $1.22 \pm 0.09$ & $68.7 \pm 2.8$ & $3.0 \pm 0.9$ \\
log-MPPI & 1 & \cellcolor{lightgray}$0.27 \pm 0.02$ & $15.2 \pm 4.6$ & $1.23 \pm 0.02$ & $63.5 \pm 1.1$ & $8.2 \pm 0.5$ \\
Ours & \cellcolor{lightgray}0 & $0.28 \pm 0.05$ & \cellcolor{lightgray}$14.4 \pm 2.2$ & \cellcolor{lightgray}$1.25 \pm 0.02$ & \cellcolor{lightgray} $63.4 \pm 1.6$ & $9.1 \pm 0.5$ \\
\hline\hline
\multicolumn{7}{|c|}{\textit{\textbf{Scenario \#3:}} \((100\%, 2)\), Non-Cooperative Mode (NCM)} \\
\hline
MPCC & 4 & $0.71 \pm 0.16$ & $60.8 \pm 12.9$ & $1.47 \pm 0.09$ & $69.1 \pm 2.4$ & $3.2 \pm 0.4$ \\
log-MPPI & 1 & $0.32 \pm 0.03$ & $18.3 \pm 6.5$ & $1.54 \pm 0.05$ & $63.7 \pm 1.3$ & $8.2 \pm 0.3$ \\
Ours & \cellcolor{lightgray}0 & \cellcolor{lightgray}$0.29 \pm 0.04$ & \cellcolor{lightgray}$17.3 \pm 2.6$ & \cellcolor{lightgray}$1.62 \pm 0.07$ & \cellcolor{lightgray}$62.8 \pm 1.4$ & $8.9 \pm 0.4$ \\
\hline
\end{tabular}\label{6-ped-NCM}
\vspace*{-8pt}
\end{table}

In Table~\ref{10-ped-NCM}, the analysis extends to the 10-pedestrian environment and includes the two U-MPPI variants, U-MPPI\textsuperscript{-1} and U-MPPI\textsuperscript{1}, which were excluded from Table~\ref{6-ped-NCM} due to minimal performance differences among sampling-based methods across all metrics, while the more complex 10-pedestrian setting reveals clearer distinctions.
Notably, both U-MPPI variants outperform MPCC and log-MPPI in achieving fewer collisions \(\mathcal{N}_{\text{c}}\) across all scenarios, while demonstrating shorter routes to the goal, particularly in \textit{Scenarios \#2} and \textit{\#3}, due to their efficient sampling strategy and integration of risk-sensitive trajectory evaluation, as described in \eqref{Qrs-modified}.
Our method consistently achieves the lowest collision rates across all scenarios, with high trajectory quality in \textit{Scenario \#1}, as reflected by shorter \(d_{\text{av}}\). However, in \textit{Scenarios \#2} and \textit{\#3}, trajectory efficiency declines, with \(d_{\text{av}}\) exceeding that of all other methods in \textit{Scenario \#2} and the U-MPPI variants in \textit{Scenario \#3}.
This reduction in efficiency results from setting \( r_r \) to $\SI{0.3}{\metre}$, rather than the default $\SI{0.2}{\metre}$, to enhance safety in high-speed scenarios. 
Our empirical observations in this crowded corridor environment revealed that increasing both \( r_r \) and \( N_p \) leads the robot to yield more space to pedestrians. For example, we observed instances where the robot occasionally moved backward to avoid collisions, particularly when approaching one of its goals such as \( G_2 \) or when encountering oncoming pedestrian groups. This behavior, as shown in the supplementary video, increased the traveled distance \( d_{\text{av}} \) while prioritizing pedestrian safety.
\begin{table}[t!]
\vspace*{-2pt}
\caption{Performance statistics of the evaluated control strategies in 10-pedestrian NCM, including U-MPPI variants (U-MPPI\textsuperscript{-1}, U-MPPI\textsuperscript{1}).}
\footnotesize\addtolength{\tabcolsep}{-5.5pt}
\setlength\extrarowheight{1pt}
\centering
\begin{tabular}{|c|c|c|c|c|c|c|}
\hline
Scheme & $\mathcal{N}_{\text{c}}$ & $E_{\text{dyn}}$ & \(10 \!\times\! J_{\text{acc}}\) [\si{\newton}] & $v_{\text{av}}$ [\si{\meter\per\second}] & $d_{\text{av}}$ [\si{\meter}] & $t_{\text{exec}}$ [\si{\milli\second}] \\
\hline\hline
\multicolumn{7}{|c|}{\textit{\textbf{Scenario \#1:}} \((30\%, 1)\), Non-Cooperative Mode (NCM)} \\
\hline
MPCC & 3 & $0.74 \pm 0.12$ & $36.8 \pm 7.3$ & \cellcolor{lightgray}$0.74 \pm 0.12$ & $72.2 \pm 5.9$ & $6.4 \pm 1.2$ \\
log-MPPI & 1 & $0.35 \pm 0.04$ & $9.6 \pm 1.9$ & $0.71 \pm 0.04$ & $76.6 \pm 3.1$ & $9.8 \pm 0.4$ \\
U-MPPI\textsuperscript{-1} & 1 & $0.36 \pm 0.04$ & $10.0 \pm 3.4$ & $0.71 \pm 0.03$ & $76.8 \pm 3.3$ & $13.1 \pm 0.5$ \\
U-MPPI\textsuperscript{1} & 1 & $0.36 \pm 0.05$ & \cellcolor{lightgray}$8.9 \pm 2.3$ & $0.70 \pm 0.05$ & $75.0 \pm 3.9$ & $12.7 \pm 0.3$ \\
Ours & \cellcolor{lightgray}0 & \cellcolor{lightgray}$0.35 \pm 0.02$ & $9.2 \pm 1.1$ & $0.72 \pm 0.04$ & \cellcolor{lightgray}$68.9 \pm 1.2$ & $11.0 \pm 0.5$ \\
\hline\hline
\multicolumn{7}{|c|}{\textit{\textbf{Scenario \#2:}} \((80\%, 1.5)\), Non-Cooperative Mode (NCM)} \\
\hline
MPCC & 9 & $0.71 \pm 0.13$ & $45.4 \pm 9.1$ & $1.10 \pm 0.10$ & $70.5 \pm 1.3$ & $6.0 \pm 1.1$ \\
log-MPPI & 4 & $0.29 \pm 0.04$ & $16.0 \pm 4.3$ & $1.19 \pm 0.06$ & $68.6 \pm 3.9$ & $9.9 \pm 0.6$ \\
U-MPPI\textsuperscript{-1} & 2 & \cellcolor{lightgray}$0.26 \pm 0.04$ & $12.8 \pm 1.5$ & \cellcolor{lightgray}$1.22 \pm 0.04$ & \cellcolor{lightgray} $67.1 \pm 3.7$ & $12.5 \pm 0.6$ \\
U-MPPI\textsuperscript{1} & 3 & $0.29 \pm 0.03$ & \cellcolor{lightgray}$12.6 \pm 0.8$ & $1.17 \pm 0.04$ & $67.7 \pm 2.8$ & $12.7 \pm 0.3$ \\
Ours & \cellcolor{lightgray}1 & $0.31 \pm 0.04$ & $17.8 \pm 4.6$ & $1.16 \pm 0.07$ & $72.4 \pm 9.3$ & $10.9 \pm 0.4$ \\
\hline\hline
\multicolumn{7}{|c|}{\textit{\textbf{Scenario \#3:}} \((100\%, 2)\), Non-Cooperative Mode (NCM)} \\
\hline
MPCC & 17 & $0.75 \pm 0.11$ & $53.3 \pm 13.4$ & $1.35 \pm 0.14$ & $71.7 \pm 3.4$ & $6.2 \pm 1.2$ \\
log-MPPI & 5 & $0.34 \pm 0.03$ & $16.6 \pm 4.3$ & $1.44 \pm 0.07$ & $69.9 \pm 5.6$ & $10.1 \pm 0.5$ \\
U-MPPI\textsuperscript{-1} & 3 & $0.34 \pm 0.07$ & $16.3 \pm 1.6$ & \cellcolor{lightgray}$1.46 \pm 0.10$ & \cellcolor{lightgray}$66.9 \pm 2.2$ & $12.7 \pm 0.4$ \\
U-MPPI\textsuperscript{1} & 4 & $0.34 \pm 0.04$ & \cellcolor{lightgray}$15.6 \pm 4.3$ & $1.42 \pm 0.09$ & $67.4 \pm 3.5$ & $12.8 \pm 0.5$ \\
Ours & \cellcolor{lightgray}0 & \cellcolor{lightgray}$0.33 \pm 0.02$ & $17.5 \pm 1.8$ & $1.44 \pm 0.09$ & $68.6 \pm 4.6$ & $11.4 \pm 0.4$ \\
\hline
\end{tabular}\label{10-ped-NCM}
\vspace*{-8pt}
\end{table}

Reducing \( N_p \) has shown promise in improving C\textsuperscript{2}U-MPPI performance, as demonstrated in Test \#1 of Table~\ref{imapct-N_p and Q}. By setting \( N_p \!=\! 1 \) instead of the default \( N_p \!=\! 2 \) for the high-interaction \textit{Scenario \#3} in the 10-pedestrian NCM setup evaluated in Table~\ref{10-ped-NCM}, we observed a high-quality trajectory with \( d_{\text{av}} = \SI{66.7}{\metre} \), compared to \SI{68.6}{\metre} with \( N_p = 2 \), along with smoother control actions. Tests \#2 and \#3 further explored performance adjustments by setting \( N_p \!=\! 0 \) (no prediction), while assessing the impact of different \( \gamma\) values, specifically, \( \gamma = 1 \) in Test \#2 and \( \gamma = -1 \) in Test \#3. We note that reducing \( N_p \) to 0 led to a decline in performance, indicating that some level of pedestrian prediction is essential to maintain trajectory efficiency and control quality.
Moreover, it is evident from Table~\ref{10-ped-NCM} that the U-MPPI variants perform similarly, due to assigning a higher value to \( Q \). To further examine this effect, Tests \#4, \#5, and \#6 replicate the simulations from \textit{Scenario \#3} with a reduced \( Q \!=\! \operatorname{Diag}(2.5, 2.5, 2) \) for log-MPPI, U-MPPI\textsuperscript{-1}, and U-MPPI\textsuperscript{1}, respectively. This reduction in \( Q \) results in slower robot convergence to the goals, causing oscillatory motion around the desired pose as the robot continually adjusts to avoid nearby agents, ultimately leading to a significantly longer trajectory. In Test \#5, however, we observe a higher-quality trajectory compared to the other two tests, as in this setting \( Q_{\text{rs}} \) increases with rising uncertainty levels, imposing greater penalties for deviations from the desired state, thereby reducing oscillations and yielding a more efficient trajectory.
\begin{table}[!t]
\vspace*{-2pt}
\caption{Impact of lower $N_p$ and reduced \( Q \) on sampling-based control strategies in a 10-pedestrian NCM setup.}
\footnotesize\addtolength{\tabcolsep}{-5pt}
\setlength\extrarowheight{1pt}
\centering
\begin{tabular}{|c|c|c|c|c|c|c|}
\hline
Test \# & $\mathcal{N}_{\text{c}}$ & $E_{\text{dyn}}$ & \(10 \!\times\! J_{\text{acc}}\) [\si{\newton}] & $v_{\text{av}}$ [\si{\meter\per\second}] & $d_{\text{av}}$ [\si{\meter}] & $t_{\text{exec}}$ [\si{\milli\second}] \\
\hline\hline
\multicolumn{7}{|c|}{\textit{\textbf{Scenario \#3:}} \((100\%, 2)\), Lower $N_p$ – C\textsuperscript{2}U-MPPI} \\
\hline
Test \#1 & 0 & $0.31 \pm 0.03$ & $16.3 \pm 1.5$ & $1.48 \pm 0.06$ & $66.7 \pm 2.6$ & $10.8 \pm 0.4$ \\
Test \#2 & 2 & $0.33 \pm 0.03$ & $17.2 \pm 5.8$ & $1.46 \pm 0.07$ & $69.6 \pm 6.3$ & $11.2 \pm 0.8$ \\
Test \#3 & 3 & $0.32 \pm 0.02$ & $15.1 \pm 1.3$ & $1.49 \pm 0.08$ & $69.9 \pm 5.6$ & $10.5 \pm 0.6$ \\
\hline\hline
\multicolumn{7}{|c|}{\textit{\textbf{Scenario \#3:}} \((100\%, 2)\), Reduced $Q$ – log-MPPI, U-MPPI\textsuperscript{±1}} \\
\hline
Test \#4 & 3 & $0.44 \pm 0.11$ & $15.7 \pm 5.7$ & $1.20 \pm 0.24$ & $74.8 \pm 5.7$ & $12.8 \pm 0.4$ \\
Test \#5 & 2 & $0.37 \pm 0.09$ & $16.5 \pm 7.3$ & $1.34 \pm 0.20$ & $70.1 \pm 6.8$ & $12.8 \pm 0.5$ \\
Test \#6 & 3 & $0.43 \pm 0.14$ & $16.6 \pm 7.1$ & $1.23 \pm 0.29$ & $73.5 \pm 11.6$ & $12.2 \pm 0.7$ \\
\hline
\end{tabular}\label{imapct-N_p and Q}
\vspace*{-8pt}
\end{table}

\subsubsection{Performance Summary and Discussion:}\label{Simulation-Summary}
Throughout all conducted simulations, MPCC exhibits higher \(E_{\text{dyn}}\) and \(J_{\text{acc}}\) due to its effort to balance path-following with collision avoidance, leading to sharp trajectory adjustments, increased energy consumption, and motion discontinuities in dynamic environments.
Additionally, all sampling-based approaches consistently achieve real-time performance, with \(t_{\text{exec}} < \SI{33.3}{\milli\second}\), regardless of whether the optimization problem involves 6 or 10 pedestrians.
In contrast, MPCC demonstrates real-time performance only when limited to 6 pedestrians (see \cite{brito2019model} for details).
Notably, our method, utilizing \( N_p = \SI{2}{\second} \), achieves execution times comparable to log-MPPI, which operates with \( N_p = \SI{0}{\second}\), owing to the dynamic obstacle representation and optimized thread utilization inherent in the U-MPPI framework (refer to \cite{mohamed2023towards} for details).
However, the U-MPPI variants exhibit higher \( t_{\text{exec}} \) values, even with \( N_p = \SI{0}{\second}\), due to their reliance on \( q_{\text{exp}} \), which imposes a greater computational burden per thread compared to the lightweight indicator function \( q_{\text{prob}} \) used in our method.
To sum up, it is not surprising that sampling-based methods outperform gradient-based approach, as they leverage a broader exploration of the control space and are inherently more robust to the nonlinearities and uncertainties present in dynamic environments. 
This enables them to generate feasible solutions while maintaining minimal constraint violations, enhancing their adaptability and efficiency in real-time navigation scenarios, even without relying on pedestrian prediction. 
\ihab{
This performance disparity is evident in the velocity constraint violation rate $\mathcal{V}_{\text{viol}}$, as defined in Section~\ref{Simulation Scenarios and Performance Metrics}: MPCC yields average violation rates between $18.7\%$ and $31.8\%$, whereas all sampling-based methods consistently remain below $2.6\%$.
Detailed violation statistics are provided in the supplementary material.
}

\subsubsection{Evaluation Against SH-MPC:}
\ihab{While MPCC serves as a baseline to highlight the strengths of sampling-based methods in dynamic settings, it does not incorporate chance constraints and thus lacks an explicit treatment of uncertainty. To enable a fair and uncertainty-aware comparison, we additionally evaluate C\textsuperscript{2}U-MPPI against Safe Horizon MPC (SH-MPC)~\cite{de2023scenario}, which enforces a joint chance constraint for risk-bounded planning.
To this end, we extended the evaluation by modifying \textit{Scenario~\#3} in the 10-pedestrian NCM, our most challenging setting, to include randomized pedestrian trajectories with varying start and goal poses; Table~\ref{sup-validation} presents the performance statistics of both methods, evaluated over 10 trials.
}
\begin{table}[t!]
\vspace*{-2pt}
\caption{Performance statistics of C\textsuperscript{2}U-MPPI and SH-MPC.}
\footnotesize\addtolength{\tabcolsep}{-5pt} 
\setlength\extrarowheight{1pt}
\centering
\begin{tabular}{|c|c|c|c|c|c|c|}
\hline
Scheme & $\mathcal{N}_{\text{c}}$ & $E_{\text{dyn}}$ & \(10 \!\times\! J_{\text{acc}}\) [\si{\newton}] & $v_{\text{av}}$ [\si{\meter\per\second}] & $d_{\text{av}}$ [\si{\meter}] & $\mathcal{V}_{\text{viol}}$ [\%]\\
\hline 
SH-MPC & 16 & \cellcolor{lightgray}$0.39 \pm 0.06$ & \cellcolor{lightgray}$31.8 \pm 3.1$ & $1.34  \pm 0.13$ & \cellcolor{lightgray}$63.9 \pm 0.8$ & $11.2 \pm 5.3$ \\
Ours & \cellcolor{lightgray}5 & $0.57 \pm 0.2$ & $44.9 \pm 4.2$ & \cellcolor{lightgray}$1.5 \pm 0.09$ & $66.1 \pm 1.9$ & \cellcolor{lightgray}$1.7 \pm 1.1$ \\
\hline
\end{tabular}
\label{sup-validation}
\vspace*{-9pt}
\end{table}
\ihab{
Notably, C\textsuperscript{2}U-MPPI results in significantly fewer collisions ($\mathcal{N}_{\text{c}} = 5$ vs.\ 16). This improvement comes with slightly higher $E_{\text{dyn}}$ and $J_{\text{acc}}$, reflecting the more assertive maneuvers required for proactive avoidance. Additionally, it maintains a higher $v_{\text{av}}$, demonstrating its ability to navigate around pedestrians rather than stopping when no feasible path is found, which leads to a longer $d_{\text{av}}$. In contrast, SH-MPC often stops in such scenarios, increasing the risk of collisions with incoming pedestrians. Furthermore, C\textsuperscript{2}U-MPPI exhibits fewer $\mathcal{V}_{\text{viol}}$, further underscoring its robustness in crowded, dynamic environments.
A supplementary simulation video is available at: \url{https://youtu.be/vn-734lBdqI}.
}

\subsection{Real-World Demonstration}\label{Real-World Demonstration}
\subsubsection{Experimental Setup and Validation Environment:}\label{Experimental Setup:real-world Environment}
The simulation setup from Section~\ref{Simulation Setup:Cluttered Environments} is adapted for experimental validation with two modifications: (i) the robot's maximum speed is set to \SI{1}{\metre\per\second}, while human agents have average maximum speeds ranging from \SI{0.6}{\metre\per\second} to \SI{1.3}{\metre\per\second}; (ii) agents' localization and tracking are obtained using a 12-camera Vicon motion capture system.
The performance evaluation is carried out in a \(\SI{7}{\metre} \times \SI{5.5}{\metre}\) rectangular workspace, populated with five pedestrians navigating between opposing corners, thereby creating collision-avoidance challenges.
The robot is tasked with navigating from the initial position \(\mathbf{x}_s = [\SI{0}{\metre}, \SI{0}{\metre}, 0^\circ]^\top\) to the target \(\mathbf{x}_f = [\SI{6.5}{\metre}, \SI{4.5}{\metre}, 50^\circ]^\top\) and returning to \(\mathbf{x}_s\).
\subsubsection{Experimental Results:}\label{Experimental results:real-world Environment}  
\begin{table}[h!]
\vspace*{-13pt}
\caption{Performance statistics of C\textsuperscript{2}U-MPPI with 5 pedestrians.}
\footnotesize\addtolength{\tabcolsep}{-5.2pt} 
\setlength\extrarowheight{1pt}
\centering
\begin{tabular}{|c|c|c|c|c|c|c|}
\hline
$N_p$ \![\si{\second}] & $\mathcal{N}_{\text{c}}$ & $E_{\text{dyn}}$ & \(10 \!\times\! J_{\text{acc}}\) [\si{\newton}] & $v_{\text{av}}$ [\si{\meter\per\second}] & $d_{\text{av}}$ [\si{\meter}] & $t_{\text{exec}}$ [\si{\milli\second}] \\
\hline 
$1$ & \cellcolor{lightgray}1 & \cellcolor{lightgray}$0.45 \pm 0.11$ & \cellcolor{lightgray}$21.4 \pm 1.6$ & \cellcolor{lightgray}$0.75 \pm 0.04$ & \cellcolor{lightgray}$20.2 \pm 2.3$ & $13.17 \pm 0.12$ \\
$0$ & 2 & $0.55 \pm 0.21$ & $29.7 \pm 4.3$ & $0.74 \pm 0.06$ & $27.4 \pm 3.2$ & $13.01 \pm 0.30$ \\
\hline
\end{tabular}
\label{expt-validation}
\vspace*{-8pt}
\end{table}
The performance statistics of C\textsuperscript{2}U-MPPI, based on six trials conducted in the indoor environment, are summarized in Table~\ref{expt-validation} for \(N_p = 1\ \mathrm{s}\) and \(N_p = 0\ \mathrm{s}\). Based on the trials, we can conclude that incorporating pedestrian predictions (\(N_p = \SI{1}{\second}\)) enhances dynamic energy efficiency \(E_{\text{dyn}}\) and control smoothness \(J_{\text{acc}}\) compared to \(N_p = \SI{0}{\second}\), with a marginal increase in average velocity \(v_{\text{av}}\) and fewer collisions \(\mathcal{N}_{\text{c}}\). However, the robot's average distance traversed \(d_{\text{av}}\) is lower, reflecting a trade-off between trajectory efficiency and safety. Execution times \(t_{\text{exec}}\) remain consistent, supporting real-time operation of the control algorithm.  
Additionally, C\textsuperscript{2}U-MPPI achieves a lower velocity violation rate \( \mathcal{V}_{\text{viol}} \), decreasing from \(6.2 \pm 2.2\%\) to \(4.3 \pm 1.7\%\) when pedestrian prediction is enabled.
\section{Conclusion}\label{sec:conclusion}

In this letter, we established that robust probabilistic collision avoidance in human-shared environments can be effectively achieved via a chance-constrained U-MPPI framework, where probabilistic collision checking is reformulated as deterministic constraints for seamless integration.
Our C\textsuperscript{2}U-MPPI framework robustly integrates predictive motion uncertainties, risk-aware constraints, and a layered dynamic obstacle representation, enabling efficient multi-obstacle handling while enhancing safety and computational efficiency.
Extensive simulations and experimental results demonstrate its superiority over baseline methods in terms of collision avoidance, trajectory efficiency, and real-time adaptability within dense crowds.
Future work will focus on extending the framework to multi-agent systems.

\bibliographystyle{IEEEtran}
\bibliography{references}

\begin{thebibliography}{10}
\providecommand{\url}[1]{#1}
\csname url@rmstyle\endcsname
\providecommand{\newblock}{\relax}
\providecommand{\bibinfo}[2]{#2}
\providecommand\BIBentrySTDinterwordspacing{\spaceskip=0pt\relax}
\providecommand\BIBentryALTinterwordstretchfactor{4}
\providecommand\BIBentryALTinterwordspacing{\spaceskip=\fontdimen2\font plus
\BIBentryALTinterwordstretchfactor\fontdimen3\font minus
  \fontdimen4\font\relax}
\providecommand\BIBforeignlanguage[2]{{%
\expandafter\ifx\csname l@#1\endcsname\relax
\typeout{** WARNING: IEEEtran.bst: No hyphenation pattern has been}%
\typeout{** loaded for the language `#1'. Using the pattern for}%
\typeout{** the default language instead.}%
\else
\language=\csname l@#1\endcsname
\fi
#2}}

\bibitem{singamaneni2024survey}
P.~T. Singamaneni, P.~Bachiller-Burgos, L.~J. Manso, A.~Garrell, A.~Sanfeliu,
  A.~Spalanzani, and R.~Alami, ``A survey on socially aware robot navigation:
  Taxonomy and future challenges,'' \emph{Int. J. Robot. Res.}, 2024.

\bibitem{ferrer2013robot}
G.~Ferrer, A.~Garrell, and A.~Sanfeliu, ``Robot companion: A social-force based
  approach with human awareness-navigation in crowded environments,'' in
  \emph{Proc. IEEE/RSJ Int. Conf. Intell. Robots Syst.}, 2013.

\bibitem{khatib1986real}
O.~Khatib, ``Real-time obstacle avoidance for manipulators and mobile robots,''
  \emph{Int. J. Robot. Res.}, vol.~5, no.~1, pp. 90--98, 1986.

\bibitem{chen2017socially}
Y.~F. Chen, M.~Everett, M.~Liu, and J.~P. How, ``Socially aware motion planning
  with deep reinforcement learning,'' in \emph{Proc. IEEE/RSJ Int. Conf.
  Intell. Robots Syst.}, 2017, pp. 1343--1350.

\bibitem{brito2019model}
B.~Brito, B.~Floor, L.~Ferranti, and J.~Alonso-Mora, ``Model predictive
  contouring control for collision avoidance in unstructured dynamic
  environments,'' \emph{IEEE Robot. Autom. Lett.}, pp. 4459--4466, 2019.

\bibitem{chen2021interactive}
Y.~Chen, F.~Zhao, and Y.~Lou, ``Interactive model predictive control for robot
  navigation in dense crowds,'' \emph{IEEE Transactions on Systems, Man, and
  Cybernetics: Systems}, vol.~52, no.~4, pp. 2289--2301, 2021.

\bibitem{stefanini2024efficient}
E.~Stefanini, L.~Palmieri, A.~Rudenko, T.~Hielscher, T.~Linder, and
  L.~Pallottino, ``Efficient context-aware model predictive control for
  human-aware navigation,'' \emph{IEEE Robot. Autom. Lett.}, 2024.

\bibitem{jian2023dynamic}
Z.~Jian, Z.~Yan, X.~Lei, Z.~Lu, B.~Lan, X.~Wang, and B.~Liang, ``Dynamic
  control barrier function-based model predictive control to safety-critical
  obstacle-avoidance of mobile robot,'' in \emph{Proc. IEEE Int. Conf. Robot.
  Automat.}, 2023, pp. 3679--3685.

\bibitem{lu2024robot}
Z.~Lu, K.~Feng, J.~Xu, H.~Chen, and Y.~Lou, ``Robot safe planning in dynamic
  environments based on model predictive control using control barrier
  function,'' \emph{arXiv preprint arXiv:2404.05952}, 2024.

\bibitem{du2011probabilistic}
N.~E. Du~Toit and J.~W. Burdick, ``Probabilistic collision checking with chance
  constraints,'' \emph{IEEE Trans. Robot.}, pp. 809--815, 2011.

\bibitem{zhu2019chance}
H.~Zhu and J.~Alonso-Mora, ``Chance-constrained collision avoidance for {MAVs}
  in dynamic environments,'' \emph{IEEE Robot. Autom. Lett.}, vol.~4, no.~2,
  pp. 776--783, 2019.

\bibitem{mohamed2023towards}
I.~S. Mohamed, J.~Xu, G.~Sukhatme, and L.~Liu, ``Towards efficient {MPPI}
  trajectory generation with unscented guidance: {U-MPPI} control strategy,''
  \emph{IEEE Trans. Robot. (T-RO)}, vol.~41, pp. 1172--1192, 2025.

\bibitem{zeng2021enhancing}
J.~Zeng, Z.~Li, and K.~Sreenath, ``Enhancing feasibility and safety of
  nonlinear model predictive control with discrete-time control barrier
  functions,'' in \emph{Proc. IEEE Conf. Decis. Control}, 2021, pp. 6137--6144.

\bibitem{ma2021feasibility}
H.~Ma, X.~Zhang, S.~E. Li, Z.~Lin, Y.~Lyu, and S.~Zheng, ``Feasibility
  enhancement of constrained receding horizon control using generalized control
  barrier function,'' in \emph{IEEE Int. Conf. Ind. Cyber-Phys. Syst.}, 2021.

\bibitem{mustafa2023probabilistic}
K.~A. Mustafa, O.~de~Groot, X.~Wang, J.~Kober, and J.~Alonso-Mora,
  ``Probabilistic risk assessment for chance-constrained collision avoidance in
  uncertain dynamic environments,'' in \emph{Proc. IEEE Int. Conf. Robot.
  Automat.}, 2023, pp. 3628--3634.

\bibitem{castillo2020real}
M.~Castillo-Lopez, P.~Ludivig, S.~A. Sajadi-Alamdari, J.~L. Sanchez-Lopez,
  M.~A. Olivares-Mendez, and H.~Voos, ``A real-time approach for
  chance-constrained motion planning with dynamic obstacles,'' \emph{IEEE
  Robot. Autom. Lett.}, vol.~5, no.~2, pp. 3620--3625, 2020.

\bibitem{ryu2024integrating}
K.~Ryu and N.~Mehr, ``Integrating predictive motion uncertainties with
  distributionally robust risk-aware control for safe robot navigation in
  crowds,'' \emph{arXiv preprint arXiv:2403.05081}, 2024.

\bibitem{mohamed2022autonomous}
I.~S. Mohamed, K.~Yin, and L.~Liu, ``Autonomous navigation of {AGVs} in unknown
  cluttered environments: {log-MPPI} control strategy,'' \emph{IEEE Robot.
  Autom. Lett.}, vol.~7, no.~4, pp. 10\,240--10\,247, 2022.

\bibitem{williams2017model}
G.~Williams, A.~Aldrich, and E.~A. Theodorou, ``Model predictive path integral
  control: From theory to parallel computation,'' \emph{Journal of Guidance,
  Control, and Dynamics}, vol.~40, no.~2, pp. 344--357, 2017.

\bibitem{thomas2022learning}
H.~Thomas, M.~G. de~Saint~Aurin, J.~Zhang, and T.~D. Barfoot, ``Learning
  spatiotemporal occupancy grid maps for lifelong navigation in dynamic
  scenes,'' in \emph{Proc. IEEE Int. Conf. Robot. Automat.}, 2022, pp.
  484--490.

\bibitem{wu2024decentralized}
S.~Wu, G.~Chen, M.~Shi, and J.~Alonso-Mora, ``Decentralized multi-agent
  trajectory planning in dynamic environments with spatiotemporal occupancy
  grid maps,'' in \emph{IEEE Int. Conf. Robot. Automat.}, 2024.

\bibitem{de2023scenario}
O.~de~Groot, L.~Ferranti, D.~M. Gavrila, and J.~Alonso-Mora, ``Scenario-based
  motion planning with bounded probability of collision,'' \emph{The
  International Journal of Robotics Research}, p. 02783649251315203, 2023.

\end{thebibliography}

\clearpage
\section*{Supplementary Material}
\label{sec:supplement}
This section provides additional implementation details and extended evaluations that support and complement the findings presented in the main article.
\subsection{Scenario Configuration and Visualization}\label{subsec:Scenario-Configuration-and-Visualization}
\begin{figure}[!h]%
    \vspace*{-8pt}
    \centering
    \subfloat[10-pedestrian corridor environment]{%
        \vspace*{3pt}\includegraphics[height=0.85in]{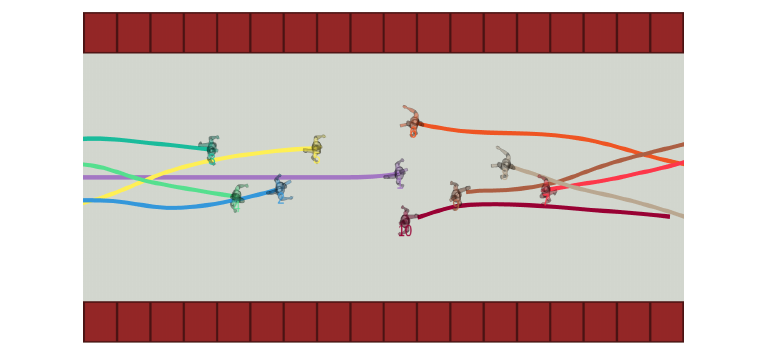}%
        \label{fig:pedsim_env}}%
    \hspace*{-0.3cm}
    \subfloat[10-pedestrian velocity profiles]{%
        \includegraphics[scale=0.99]{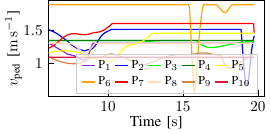}%
        \label{fig:ped-velocities}}%
    \vspace*{-3pt}
    \caption{Snapshot of (a) our simulated corridor environment with 10 pedestrians moving at different speeds along the corridor, and (b) the velocity profiles of each pedestrian at 100\% of their maximum speed.}
    \label{fig:pedsim-env-example}%
    \vspace*{-14pt}
\end{figure}

Figure~\ref{fig:pedsim-env-example} provides additional insights into our simulated corridor environment used for performance evaluation, illustrating the 10-pedestrian configuration. As mentioned in the main text, we utilize a \(\SI{30}{\metre} \times \SI{6}{\metre}\) corridor environment featuring pedestrian groups of either 6 or 10 individuals. 
As illustrated in Fig.~\ref{fig:pedsim_env}, each configuration consists of two pedestrian groups moving in opposing directions, with time-varying walking speeds introduced to emulate realistic interactions. The mean walking speeds, computed over individual trajectories, range from \SI{1.1}{\metre\per\second} to \SI{1.9}{\metre\per\second}, as depicted in Fig.~\ref{fig:ped-velocities}.
Pedestrians' trajectories are strategically designed to ensure even spatial distribution across the environment and meaningful intersections with the robot’s target poses, thereby increasing scenario complexity. In the 10-pedestrian setup, for example, at least four individuals cross or approach the robot’s destinations, enabling dynamic interaction and collision avoidance while testing the robot’s response to oncoming pedestrians when stationary.

\subsection{Extended Simulation Results}
\label{subsec:exp_results}
\subsubsection{Evaluating 2D Costmap Navigation in Dynamic Scenes:}
In Section~\ref{Cost Function Design}, we stated that using only the first two terms of the cost function \( q(\cdot) \) in \eqref{eq:traj-cost} does not adequately enhance safety when relying solely on a 2D local costmap. To support this, we conducted additional simulations with log-MPPI and U-MPPI\textsuperscript{1} in \textit{Scenario~\#3} of the 10-pedestrian NCM setup, using the 2D costmap to avoid both static and dynamic obstacles. Log-MPPI and U-MPPI\textsuperscript{1} resulted in 20 and 13 collisions, respectively, underscoring the limitations of costmap-only navigation and the need for the proposed dynamic cost \( q_{\text{dyn}} \) to improve safety.

\subsubsection{Performance in Cooperative Pedestrian Environments:}
\ihab{In cooperative mode (CM), pedestrians are modeled to actively avoid the robot, resulting in fewer conflicts and a lower risk of collision. While this behavior reflects real-world social navigation, it inherently reduces the complexity of the planning problem, thereby limiting the rigor with which a controller’s robustness can be evaluated.
To better illustrate this limitation and provide additional insight, we extend the analysis to include 10-trial simulations comparing the gradient-based MPCC and the sampling-based log-MPPI under CM, as presented in Table~\ref{mppc-logmppi-10ped-CM}.
We exclude C\textsuperscript{2}U-MPPI from this comparison, as log-MPPI represents a worst-case scenario for sampling-based methods without pedestrian prediction (\(N_p = \SI{0}{\second}\)).
Notably, both control strategies achieve collision-free navigation across all scenarios; however, log-MPPI demonstrates enhanced efficiency and smoothness, reflected in its lower averages in accumulated jerk \( J_{\text{acc}} \), distance traveled \( d_{\text{av}} \), and dynamic energy \( E_{\text{dyn}} \), alongside a higher average speed \( v_{\text{av}} \).
This contrasts with the results in Table~\ref{10-ped-NCM}, where the same setup under non-cooperative mode (NCM) leads to noticeable performance degradation due to more complex and uncooperative pedestrian interactions.
Therefore, the primary analysis in Section~\ref{Simulation-Results} focuses on NCM, where pedestrians move independently, challenging the control strategy's collision avoidance capabilities and adaptability under uncooperative conditions.}
\begin{table}[!t]
\caption{MPCC and log-MPPI performance statistics across ten trials in a cooperative mode (CM) with 10 pedestrians.}
\footnotesize\addtolength{\tabcolsep}{-5.2pt} 
\setlength\extrarowheight{1pt}
\centering
\begin{tabular}{|c|c|c|c|c|c|c|}
\hline
Scheme & $\mathcal{N}_{\text{c}}$ & $E_{\text{dyn}}$ & \(10 \!\times\! J_{\text{acc}}\!\) [\si{\newton}] & $v_{\text{av}}$ [\si{\meter\per\second}] & $d_{\text{av}}$ [\si{\meter}] & $t_{\text{exec}}$ [\si{\milli\second}] \\
\hline \hline
\multicolumn{7}{|c|}{\textit{\textbf{Scenario \#1:}} \((30\%, 1)\), Cooperative Mode (CM)} \\
\hline
MPCC & \cellcolor{lightgray}0 & $0.73 \pm 0.06$ & $32.9 \pm 4.8$ & $0.73 \pm 0.10$ & $67.1 \pm 1.5$ & $5.3 \pm 0.6$ \\
log-MPPI & \cellcolor{lightgray}0 & \cellcolor{lightgray}$0.34 \pm 0.07$ & \cellcolor{lightgray}$9.6 \pm 2.7$ & \cellcolor{lightgray}$0.77 \pm 0.07$ & \cellcolor{lightgray}$66.4 \pm 3.7$ & $9.8 \pm 0.4$ \\
\hline\hline
\multicolumn{7}{|c|}{\textit{\textbf{Scenario \#2:}} \((80\%, 1.5)\), Cooperative Mode (CM)} \\
\hline
MPCC & \cellcolor{lightgray}0 & $0.74 \pm 0.07$ & $51.5 \pm 4.2$ & $1.22 \pm 0.08$ & $67.0 \pm 2.2$ & $4.3 \pm 0.8$ \\
log-MPPI & \cellcolor{lightgray}0 & \cellcolor{lightgray}$0.26 \pm 0.02$ & \cellcolor{lightgray}$13.7 \pm 2.3$ & \cellcolor{lightgray}$1.23 \pm 0.03$ & \cellcolor{lightgray}$62.5 \pm 0.7$ & $8.9 \pm 0.5$ \\
\hline\hline
\multicolumn{7}{|c|}{\textit{\textbf{Scenario \#3:}} \((100\%, 2)\), Cooperative Mode (CM)} \\
\hline
MPCC & \cellcolor{lightgray}0 & $0.74 \pm 0.03$ & $58.7 \pm 8.0$ & $1.46 \pm 0.17$ & $66.6 \pm 2.8$ & $4.8 \pm 1.7$ \\
log-MPPI & \cellcolor{lightgray}0 & \cellcolor{lightgray} $0.32 \pm 0.04$ & \cellcolor{lightgray}$16.4 \pm 1.4$ & \cellcolor{lightgray}$1.56 \pm 0.05$ & \cellcolor{lightgray}$63.3 \pm 1.4$ & $9.1 \pm 0.5$ \\
\hline
\end{tabular}
\label{mppc-logmppi-10ped-CM}
\vspace*{-8pt}
\end{table}

\subsubsection{Discussion on Velocity Constraint Violations:} 
As evidenced by the results in Section~\ref{Simulation-Results}, the sampling-based methods, including log-MPPI, U-MPPI variants, and C\textsuperscript{2}U-MPPI, demonstrate enhanced performance in terms of collision avoidance, control smoothness, and constraint handling, even when operating with \( N_p = 0\,\text{s} \). To further evaluate how well these methods adhere to control constraints, Table~\ref{violation-anaysis} reports the average violation rate \( \mathcal{V}_{\text{viol}} \) of the robot’s desired linear velocity across all scenarios for both cooperative (CM) and non-cooperative (NCM) settings, as presented in Tables~\ref{mppc-logmppi-10ped-CM},~\ref{6-ped-NCM}, and~\ref{10-ped-NCM}. 
The results show that MPCC consistently yields the highest violation rates, often reaching or exceeding $18.7\%$ on average, with some scenarios approaching or surpassing $30\%$. In contrast, all sampling-based methods maintain substantially lower $\mathcal{V}_{\text{viol}}$ values, generally remaining below $3.5\%$ even when accounting for variability, demonstrating their robustness in generating dynamically feasible and constraint-aware behaviors.
The last column in Table~\ref{violation-anaysis} presents the metric \( \mathcal{S}_{\text{inf}} \), defined as the average percentage of time steps at which the MPCC solver fails to return a feasible solution. These solver infeasibilities directly contribute to the elevated velocity constraint violations observed in MPCC, particularly in pedestrian-dense or highly interactive scenarios.
\begin{table}[h!]
\vspace*{-8pt}
\caption{Average velocity constraint violation \( \mathcal{V}_{\text{viol}} \) for all methods, and solver infeasibility \( \mathcal{S}_{\text{inf}} \) for MPCC only, across all scenarios.}
\footnotesize\addtolength{\tabcolsep}{-4.2pt} 
\setlength\extrarowheight{1pt}
\centering
\begin{tabular}{|c|c|c|c|c|c|c|}
\hline
Scen. & MPCC & log-MPPI & U-MPPI\textsuperscript{-1} & U-MPPI\textsuperscript{1} & C\textsuperscript{2}U-MPPI & \( \mathcal{S}_{\text{inf}} \) [\%]  \\
\hline\hline
\multicolumn{7}{|c|}{\textbf{\textit{Cooperative Mode (CM):}} 10-pedestrian setup (Table~\ref{mppc-logmppi-10ped-CM})} \\
\hline
\#1 & $26.9 \pm 4.2$ & $1.4 \pm 0.6$ & $-$ & $-$ & $-$ & $3.8 \pm 3.2$\\
\#2 & $18.7 \pm 8.1$ & $0.5 \pm 0.3$ & $-$ & $-$ & $-$ & $6.7 \pm 3.9$\\
\#3 & $22.3 \pm 4.9$ & $0.4 \pm 0.6$ & $-$ & $-$ & $-$ & $8.5 \pm 6.1$\\
\hline\hline
\multicolumn{7}{|c|}{\textbf{\textit{Non-Cooperative Mode (NCM):}} 6-pedestrian setup (Table~\ref{6-ped-NCM})} \\
\hline
\#1 & $30.0 \pm 3.3$ & $2.0 \pm 1.0$ & $-$ & $-$ & $2.6 \pm 0.8$ & $1.4 \pm 2.0$\\
\#2 & $25.1 \pm 8.4$ & $1.1 \pm 0.9$ & $-$ & $-$ & $0.7 \pm 0.5$ & $8.0 \pm 5.1$\\
\#3 & $25.7 \pm 7.8$ & $0.4 \pm 0.5$ & $-$ & $-$ & $0.7 \pm 0.8$ & $10.9 \pm 4.9$\\
\hline\hline
\multicolumn{7}{|c|}{\textbf{\textit{Non-Cooperative Mode (NCM):}} 10-pedestrian setup (Table~\ref{10-ped-NCM})} \\
\hline
\#1 & $31.8 \pm 5.8$ & $1.6 \pm 0.7$ & $2.4 \pm 0.8$ & $2.0 \pm 0.6$ & $2.3 \pm 0.4$ & $5.8 \pm 5.6$\\
\#2 & $29.2 \pm 6.6$ & $1.9 \pm 1.2$ & $1.7 \pm 0.7$ & $1.9 \pm 0.9$ & $1.2 \pm 0.6$ & $11.0 \pm 4.7$\\
\#3 & $24.6 \pm 5.5$ & $1.4 \pm 0.7$ & $0.7 \pm 0.7$ & $0.8 \pm 0.6$ & $0.3 \pm 0.2$ & $15.3 \pm 5.6$\\
\hline
\end{tabular}\label{violation-anaysis}
\vspace*{-8pt}
\end{table}

\end{document}